\documentclass[mathpazo]{cicp}

\usepackage{amssymb, amsmath, amsthm, bm}
\usepackage[noend]{algpseudocode}
\usepackage{algorithmicx,algorithm}
\usepackage{graphicx}
\usepackage{subfig}
\usepackage{caption}
\usepackage[colorlinks,linkcolor=blue]{hyperref}
\usepackage{epstopdf}
\usepackage[colorlinks,linkcolor=blue]{hyperref}
\usepackage{bbding}
\usepackage{booktabs}
\usepackage{tikz,xcolor}
\newtheorem{rmrk}{Remark}

\begin{document}
\title{Less Emphasis on Hard Regions:
 Curriculum Learning of PINNs for Singularly Perturbed Convection-Diffusion-Reaction Problems}


\author[Wang et.~al.]{Yufeng Wang \affil{1},
      Cong Xu \affil{2}, Min Yang \affil{1} \comma\corrauth and Jin Zhang\affil{3}}
\address{\affilnum{1}\ School of Mathematics and Information Sciences, Yantai University, Yantai, China \\
               \affilnum{2}\ School of Computer Science and Technology, East China Normal University, Shanghai, China \\
               \affilnum{3}\ Department of Mathematics, Shandong Normal University, Jinan, China}
 \emails{{\tt zytuyufengwang@163.com} (Y.~Wang), {\tt congxueric@gmail.com} (C.~Xu),
        {\tt yang@ytu.edu.cn} (M.~Yang),{\tt jinzhangalex@hotmail.com} (J.~Zhang)}
\begin{abstract}
      Although Physics-Informed Neural Networks (PINNs) have been successfully applied in a wide variety of science and engineering fields,
      they can fail to accurately predict the underlying solution in slightly challenging convection-diffusion-reaction problems.
      In this paper, we investigate the reason of this failure from a domain distribution perspective,
      and identify that learning multi-scale fields simultaneously makes the network unable to advance its training and easily get stuck in poor local minima.
      We show that the widespread experience of sampling more collocation points in high-loss layer regions hardly help optimize and may even worsen the results.
      These  findings motivate the development of a novel curriculum learning method that encourages neural networks to prioritize learning on easier non-layer regions
      while downplaying learning on harder layer regions.
      The proposed method helps PINNs automatically adjust  the learning emphasis and thereby facilitate the optimization procedure.
      Numerical results on typical benchmark equations show that the proposed curriculum learning approach
      mitigates the failure modes of PINNs and can produce accurate results for very sharp boundary and interior layers.
      Our work reveals that for equations whose solutions have large scale differences,
      paying less attention to high-loss regions can be an effective strategy for learning them accurately.
\end{abstract}


\maketitle

\section{Introduction}
Convection-diffusion-reaction problems appear in the modeling of various modern complicated processes,
such as fluid flow at high Reynolds numbers \cite{Hirsch2007},
drift diffusion in semiconductor device modeling \cite{Polak1987},
and chemical reactor theory \cite{Miller1997}.
Very often the size of diffusion is characterized by a parameter $\epsilon$,
which could be smaller by several orders of magnitude compared to the size of convection and/or reaction,
resulting narrow boundary or interior layers  in which the solution changes extremely rapidly \cite{Roos2008}.
Classical numerical methods use layer-adapted meshes or introduce carefully designed artificial stability terms to solve these challenging problems
\cite{Ayuso2009,Brooks1980,Stynes2003,Zhang2016,ZhangLiu2016}.

In recent years, there has been a surge of interest in applying neural networks in traditional scientific modeling (e.g. partial differential equations),
which yields the so-called physics-informed neural networks (PINNs) \cite{Col2021,Guo2022,Han2018,Huang2022,
Kris2021,Lu2021,Raissi2019,WangYu2021}.
The main idea of PINNs is to include physical domain knowledge as soft constraints in the empirical loss function and
then use existing machine learning methodologies such as stochastic optimization, to train the model.
As an interesting alternative to traditional numerical solvers,
PINN has the advantage of flexibility in dealing with high-dimensional PDEs in complicated geometry and easy incorporation of available data information.
Moreover, well-trained PINNs can have good generalization ability and can quickly predict solutions outside the computational area.

However, as reflected in some recent studies on the "failure modes" of PINNs \cite{Arz2023,Daw2022,Kris2021},
it has been found that PINNs can fail to converge to the correct solution  even for relatively simple convection-diffusion problems.
Approaches to improve the accuracy of PINNs in solving convection-diffusion problems can be broadly classified into two categories.
The first category borrows theories and concepts from conventional numerical methods.
For example,
Mojgani et al. \cite{Moj2023} rewrote the original equation into a Lagrangian form on the characteristic curves and then applied a two-branch neural network to solve the reformulated form.
However, the approach is only applicable to time-dependent problems and not to steady-state equations.
Recently, inspired by the theory of singular perturbation and asymptotic expansions,
Arzani et al. \cite{Arz2023} used separate neural networks to learn the different levels on the inner and outer layer regions, respectively.
The second category emphasizes machine learning techniques, such as the design of loss functions, sample selection, and learning strategies.
He et al. \cite{He2021} used a weighted sum of residual losses and showed that  in order to obtain an accurate solution of the advection-dispersion equation,
 the weights of the initial and boundary conditions should be larger than the PDE residuals.
Daw et al. \cite{Daw2022} proposed an evolutionary sampling algorithm in which the collocation points evolve gradually with training to prioritize high-loss regions
while maintaining a background distribution of uniformly sampled points.
Krishnapriyan et al. \cite{Kris2021}  argued that the PDE-based soft constraints make the loss landscapes difficult to optimize,
and proposed a curriculum approach that sets the PINN loss term starting with a simple equation regularization
and progressively become more complex as the network gets trained.
But for strong singular perturbation problems, the approach can be computationally overburdened due to the need to learn many intermediate subproblems.

The existing studies mainly considered the relatively simple cases where the viscosity/diffusivity is about a scale of $ 10^{-4} $.
Singularly perturbed problems containing extremely sharp layers (strong vanishing viscosity/diffusivity limit) remains an urgent target for PINNs.
This paper aims to unravel the failure modes of PINNs from some new perspectives and to further advance the approximation performance of PINNs.
We show that simultaneously learning multi-scale solutions in layer and non-layer regions makes the network difficult to advance its training and easily get stuck in poor local minima.
We demonstrate that in such case,
prioritizing layer regions (sampling more collocation points in high-loss regions)  can make the training more difficult and worsen the performance.
This surprising finding is contrary to the majority of existing studies on PINNs.
While most previous studies have emphasized high-loss regions,
our investigation indicates that for problems containing samples with extreme scale differences,
it seems not a good idea to emphasize high-loss regions.
We argue that this is because  collocation points from layer regions are significantly more challenging to learn than those from non-layer regions.
To alleviate the learning difficulties,
we propose a novel curriculum learning approach that can automatically adjust the sample weights to
emphasize easier non-layer regions,
thereby improving  the approximation accuracy of the network for strongly singular perturbation problems.
We empirically demonstrate the efficiency of  the proposed approach in a variety of typical convection-diffusion-reaction problems.
We show that the proposed curriculum learning algorithm can mitigate the failure modes of  vanilla PINNs
and well capture the sharp boundary or interior layers even in  the cases of very small diffusivity ($\epsilon= 10^{-9} $).
Our approach successfully learns solutions containing very sharp layers, using only one neural network, without learning any intermediate solutions.
More importantly,  we provide a new perspective to understand the failure modes of PINNs and reveal that for equations whose solutions have large scale differences,
paying less attention to high-loss regions could be a feasible strategy for learning them accurately.
The source code built on PyTorch is available at \url{https://github.com/WYu-Feng/CLPINN} to enable other researchers to reproduce and extend the results.

The remainder of the paper is organized as follows.
Section \ref{section-preliminaries} gives the problem under study and introduces the basic notation of PINNs.
A toy example is used in Section \ref{section-difficult}  to explore the possible reason for the failure mode of PINNs in solving singularly perturbed equations.
In Section \ref{section-ourapproach}, we design a curriculum learning approach to improve the performance of PINNs.
Section \ref{section-exp} gives comprehensive  experimental results to demonstrate the efficiency of the proposed method.
Finally, the conclusion is drawn in Section \ref{con}.

\section{Problem Setup}
\label{section-preliminaries}
Consider the following singularly perturbed equation:
\begin{align}
\label{sPDE}
   \mathcal{L} u:=\epsilon \mathcal{L}_2 u+\mathcal{L}_1 u+\mathcal{L}_0 u=f(\bm{x}),
   \quad
   \bm{x}\in \Omega,
\end{align}
where $ \Omega $ is a physical domain in $\mathbb{R}^d$,
$\mathcal{L}_k $ represents the differential operator of order  $k $, $k=0,1,2$,
$ f(\bm{x}) $ denotes the source term,
and the diffusion coefficient satisfies  $  0 < \epsilon \leq 1$.
Further assume that the solution $u(\bm{x}) $ satisfies the following boundary condition
\begin{align}
\label{Boun}
   \mathcal{B} u=g(\bm{x}), \quad \bm{x}\in \partial \Omega.
\end{align}
where $\mathcal{B}$ is a well-defined differential operator for determining the condition on the admissible boundary $\partial \Omega $.
When the diffusion coefficient $\epsilon$ is very small,
the latent solution of the equation changes rapidly within some thin layers,
posing a great challenge to the numerical simulation \cite{Brooks1980,Miller1996}.

For PINNs, the solution $u(\bm{x}) $ is approximated by a neural network  $ u_\theta(\bm{x}) $,
where $ \theta $ denotes the parameters of the network.
Let
\begin{align}
\label{ploss}
   L_{phys}(\theta)= \frac{1}{N} \sum_{i=1}^N r_{phys}^2(\bm{x}_i;\theta)= \frac{1}{N} \sum_{i=1}^N [\mathcal{L} u_\theta(\bm{x}_i)-f(\bm{x}_i)]^2
\end{align}
be the mean-squared physical residual loss of $ N $ training sample points in $\Omega $,
and
\begin{align}
\label{bloss}
   L_{bc}(\theta)= \frac{1}{M} \sum_{i=1}^M r_{bc}^2(\bm{x}_i;\theta)=\frac{1}{M} \sum_{i=1}^M [\mathcal{B} u_\theta(\bm{x}_i)-g(\bm{x}_i)]^2
\end{align}
be  the mean-squared boundary loss of $ M $  training sample points on $\partial \Omega $.
All the samples constitute a training set $ X_{train} $.

The neural network approximation $  u_\theta(\bm{x}) $  can be determined by solving  the following optimization objective
\begin{align}
\label{resloss}
  \min_\theta L_{phys}(\theta)+ \lambda L_{bc}(\theta),
\end{align}
where $\lambda $ is a hyperparameter to balance the weights of the two loss terms.

Although PINNs have been successfully applied in solving many types of differential equations,
their performance for relatively simple convection-diffusion equations are far from satisfactory.
In the next section, we  are to analyze the dilemma encountered by PINNs.

\section{Analysis of Failure Mode}
\label{section-difficult}

Consider the following one-dimensional convection-diffusion problem as an example:
\begin{align}
\label{toyEqn}
\begin{split}
   -\epsilon u_{xx}+(x-2)u_x &=f(x), \quad x\in (0,1),
   \\[5pt]
   u(0)=u(1)&=0.
\end{split}
\end{align}
where the diffusion coefficient $\epsilon $ is set as $ 10^{-3} $, and the source term $ f(x) $ is determined by the exact solution $u(x)=\cos(\pi x/2)(1-\exp(2x/\epsilon)) $.
This problem has a boundary layer at $ x=0 $.

Consider a four-layer fully connected neural network $u_{\theta}(x)$,
where each intermediate layer has 20 neurons and Tanh is used as the activation function.
The training set $X_{train} $ consists of 2500 points uniformly sampled from the domain $ (0,1) $.

\textbf{Different initializations and optimizations.}
The network parameters are initialized by Normal Xaiver or Uniform Xaiver methods \cite{Glorot2010}.
Two mainstream optimizers, Stochastic Gradient Descent (SGD) \cite{Bottou1991} and Adam \cite{King2015},
are utilized to solve the optimization objective \eqref{resloss},
where the balance parameter $ \lambda $ is set to 1.

\begin{figure}[hbt]
\centering
\subfloat[SGD]{
    \includegraphics[width=0.35\linewidth]{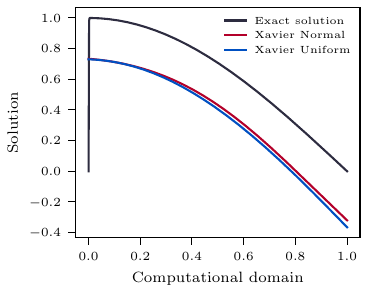}
    \label{fig-solution-sgd}
}
\subfloat[Adam]{
    \includegraphics[width=0.35\linewidth]{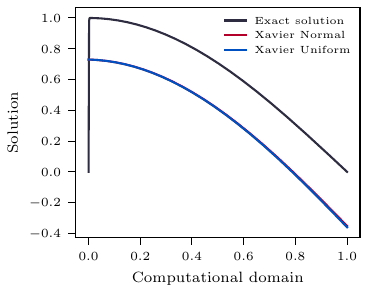}
    \label{fig-solution-adam}
}
\caption{Predictions of PINN under two parameter initializations using SGD and Adam optimizers, respectively.}
\label{NNsToy}
\end{figure}

\begin{figure}[hbt]
\centering
\subfloat[SGD]{
    \includegraphics[width=0.35\linewidth]{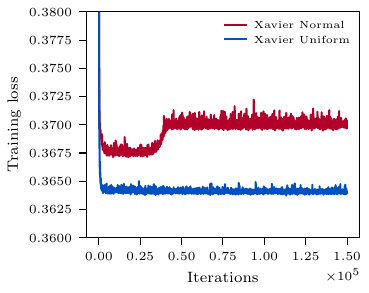}
    \label{fig-loss-sgd}
}
\subfloat[Adam]{
    \includegraphics[width=0.35\linewidth]{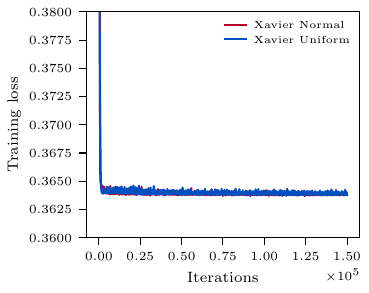}
    \label{fig-loss-adam}
}
\caption{Training loss curves of PINN under various parameter initializations using SGD and Adam optimizations, respectively.}
\label{TrainToy}
\end{figure}

\begin{figure}[hbt]
\centering
    \includegraphics[width=0.4\linewidth]{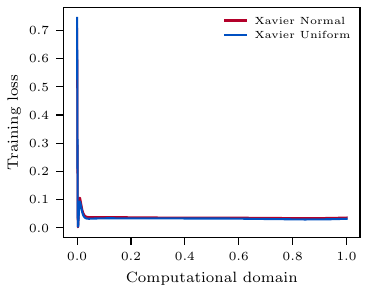}
\caption{Training loss distribution of equation \eqref{toyEqn} under different initializations using the Adam optimizer. }
\label{DomLoss}
\end{figure}

It can be observed from Figure \ref{NNsToy} that the prediction $u_{\theta}(x)$ has very large errors
throughout the computational domain, regardless of the initial or training methods used.
When we further plot the corresponding training loss curves (Figure \ref{TrainToy}),
it is clear that the training loss of PINN fails to converge even after very long iterations.
In particular, it can be seen that the training losses in the layer regions are much higher than those in the non-layer regions  (Figure \ref{DomLoss}).

\begin{figure}[!hbt]
\centering
    \scalebox{.35}{\includegraphics{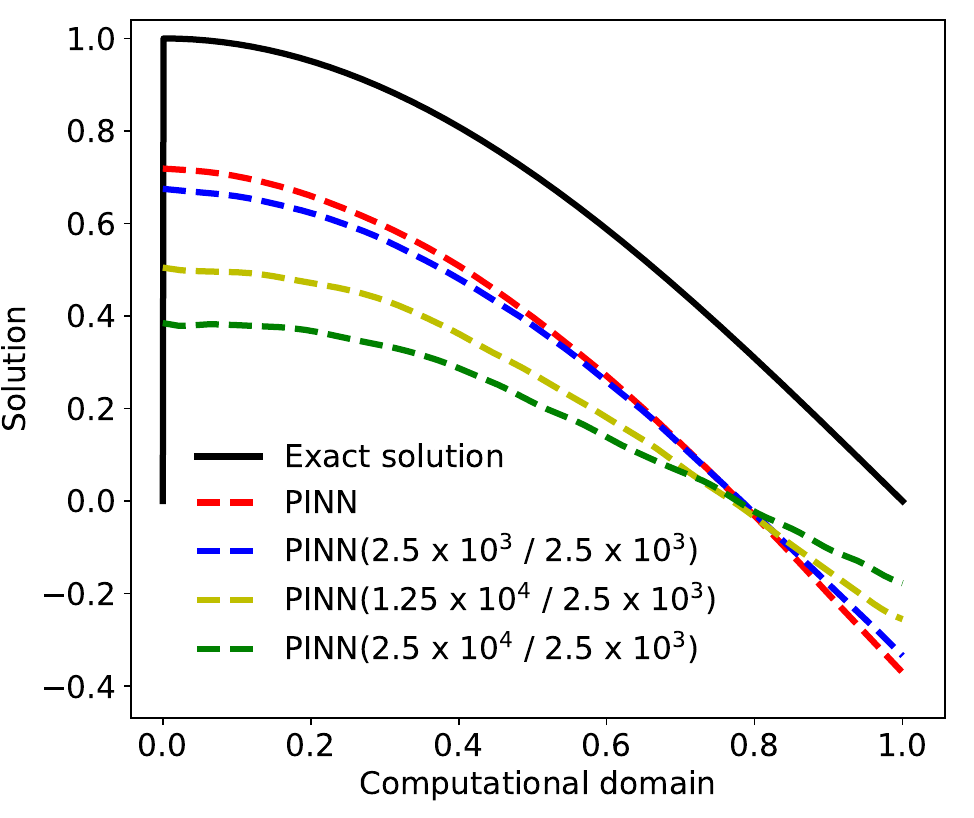}}
\caption{Predictions of PINN using a dense  sampling in the layer domain,
where 2500 points are sampled in  the  non-layer domain $ (0.1,1) $,
and 2500, 12500, 25000 points are sampled in $ (0, 0.1)$, respectively.
For standard PINN, we apply a random sampling in $ (0,1) $.
}
\label{Encry}
\end{figure}

\textbf{Emphasizing high-loss layer regions?}
Note that there exists a widely accepted consensus that the performance of PINNs can be improved by sampling more collocation points in high-loss regions.
We tried such a strategy, but unfortunately it can be found from Figure \ref{Encry} that instead of improving the approximations,
the dense sampling in the high-loss layer region may lead to worse results.

The above experiments show that for singular perturbation equations,
common PINNs cannot solve them well even with dense sampling in the high-loss layer regions.
Such paradoxical phenomenon leads to the natural question of what is the cause of this undesirable performance.

\textbf{Less emphasis on layer regions.}
We notice that compared with ordinary equations,
the latent solutions of singular perturbation equations exhibit sharp scale variations in different regions.
In the narrow layer region the solution transits very rapidly,
while in the wide non-layer region the solution varies more flatly and slowly.
We argue that such large scale differences make PINNs difficult to balance the learning of collocation points from the layer and non-layer regions.
The final loss distribution in Figure \ref{DomLoss} shows the training losses for samples close to the boundary layer are much larger than those in the non-layer domain,
which implies that the sharp layer domain may be too difficult for PINNs to learn.

\begin{figure}[hbt]
\centering
\subfloat[]{
    \includegraphics[width=0.35\linewidth]{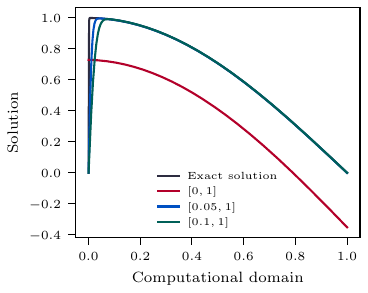}
    \label{fig-solution-interval}
}
\subfloat[]{
    \includegraphics[width=0.35\linewidth]{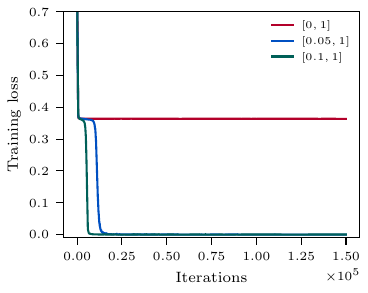}
    \label{fig-loss-interval}
}
\caption{ (a) Predictions of PINN after a rejection of the layer samples.
 (b) The corresponding training loss curves.}
 \label{Ignor}
\end{figure}

In order to reduce the learning difficulty of PINN, we put forward the following experiment.
We only select samples from non-layer regions to build the training composed of samples in $(a,1)$,
where $a$ is set to 0.05 and 0.1, respectively.
It is surprising to observe from Figure \ref{Ignor} that such a brutal discarding of layer samples can result an obvious improvement for the prediction of PINN.
Thereby, the above attempt inspires us that ``less emphasis on layer regions" may help to raise the performance of PINNs in solving singularly perturbed problems.

Of course, naively rejecting samples from the layer regions will inevitably result in the loss of important physical information,
thus cannot guarantee the high accuracy of the prediction.
Moreover, the location of the layers is usually not known in practice.
Therefore, in the next section, we are to present a curriculum learning algorithm
that dynamically estimates the location of layers and adaptively adjusts the importance of the samples close to the layers.

\begin{figure}[hbt]
\centering
   \includegraphics[width=1\linewidth]{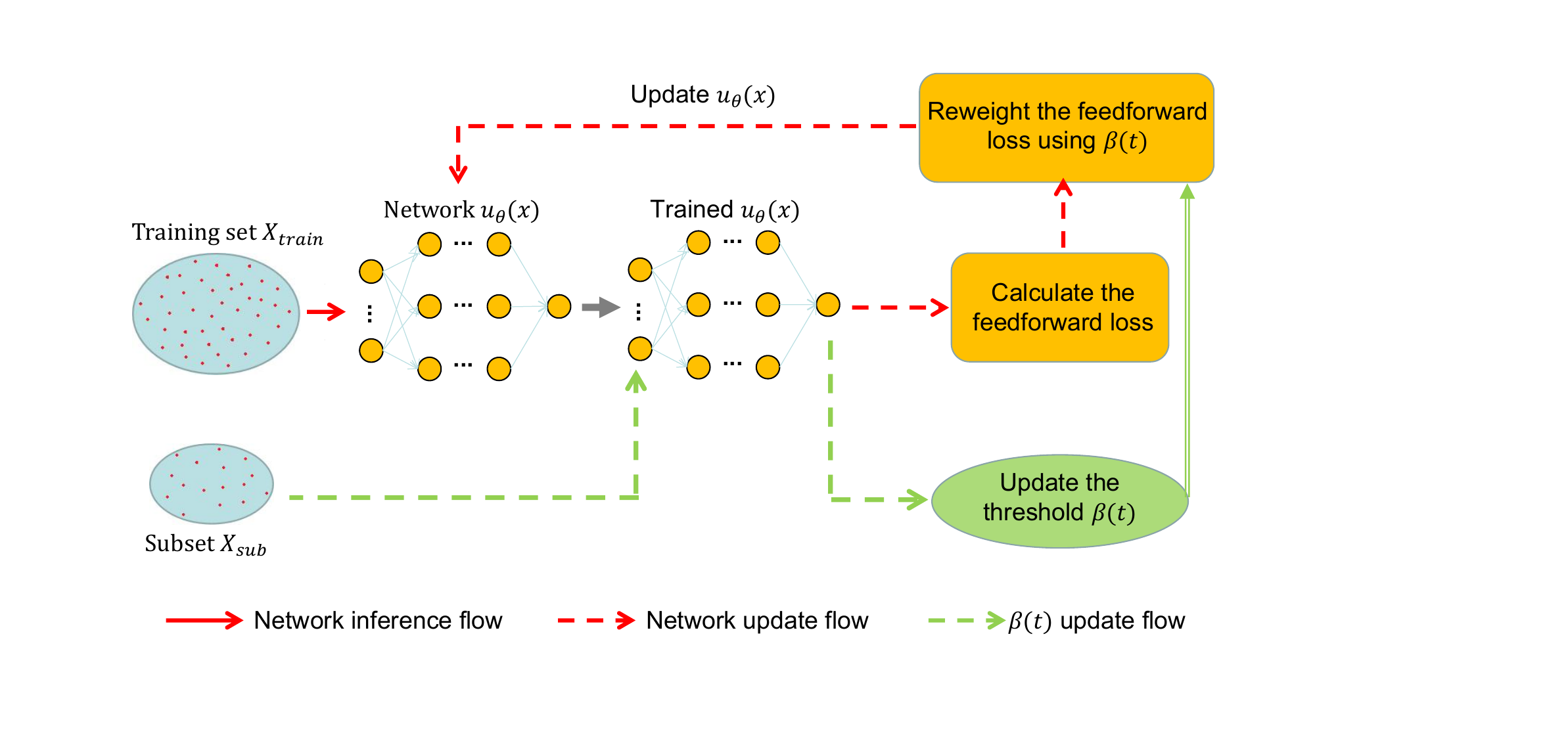}
   \caption{The proposed learning framework. A sub-training set $ X_{sub} $ is employed to update the threshold $\beta(t)$ at the $t$-th iteration step.
   Then, the threshold $\beta(t)$ is used to dynamically re-weight each training sample,
   especially lightening  the importance of the samples close to the layer regions.}
 \label{frameworkfig}
\end{figure}

\section{The Proposed Curriculum Learning}
\label{section-ourapproach}
So far, we have demonstrated that the failure mode of PINN is due to large discrepancy in sample difficulties between layer and non-layer regions.
In this section, we are to provide a curriculum learning algorithm that encourages the network to prioritize learning easier non-layer regions.

\subsection{Surrogate for layer location}
\label{Proxy}
Since the learning difficulty in layer and non-layer regions differs significantly,
the first key step is to estimate the location of the layers.
According to Figure \ref{DomLoss},
it can be found that  that the layer region usually corresponds to  a larger training loss.
Therefore, we can take the feedforward training loss as a proxy to estimate the location of the layers.
A larger loss implies that the corresponding sample is closer to the layer.

\subsection{Importance reweighting}
\label{reweight}
Recall that the the optimization objective \eqref{ploss} is the average of the squared losses of all samples:
\begin{align*}
   L_{phys}(\theta)= \frac{1}{N} \sum_{i=1}^N r_{phys}^2(\bm{x}_i;\theta),
\end{align*}
which means that samples from different regions are of equal importance for learning.

In order to make PINN place less emphasis on the samples from the layer regions,
we modify the optimization objective as follows
\begin{align}
\label{Mploss}
   L_{phys}(\theta)= \frac{1}{\sum_{i=1}^N w(\bm{x}_i)} \sum_{i=1}^N w(\bm{x}_i) r_{phys}^2(\bm{x}_i;\theta),
\end{align}
where $w(\bm{x}_i) $ represents the importance of the sample.
The closer the sample is to the layer, the less weight it has.

As discussed in Section \ref{Proxy},
we do not known the exact locations of the layers
and shall estimate them using the training losses that vary dynamically  with iterations.
Therefore, the weight of each sample should also be dynamically adjusted.
To this end, we define
\begin{align}
\label{Mploss}
   L_{phys}(\theta)= \frac{1}{\sum_{i=1}^N w(t,\bm{x}_i)} \sum_{i=1}^N w(t,\bm{x}_i) r_{phys}^2(\bm{x}_i;\theta),
\end{align}
where $t $ denotes the iteration step.
The sample weights in \eqref{Mploss} can be determined by
\begin{align}
\label{SampleWeight}
\begin{split}
       w(t,\bm{x}_i)&=\left\{\begin{array}{ll}
                            1, & \text{if}\; r_{phys}^2(\bm{x}_i)\leq \beta(t), \\[5pt]
                            \displaystyle\frac{\beta(t)}{r_{phys}^2(\bm{x}_i)}, & \text{if} \;  r_{phys}^2(\bm{x}_i)>\beta(t),
                      \end{array}\right.
\end{split}
\end{align}
where $\beta(t)$ is a loss threshold  to be updated adaptively with iterations.

Intuitively, the formula \eqref{SampleWeight} indicates that  if the training loss  of a sample is greater than $\beta(t)$,
which implies that the collocation point is close to the layer,
then we give this sample a weight $ \beta(t)/r_{phys}^2(\bm{x}) $, which is less than 1.
The larger the loss,
the closer the sample is to the layer,
and the smaller the corresponding weight.
In this way, we not only emphasize the learning of easy non-layer region samples,
but also maintain the necessary physical information of the high-loss layer regions.

\subsection{Calculate the threshold by a sub-training set}
\label{DetThr}
Since the training loss of a desirable network model will gradually descent with iterations,
then the threshold $\beta(t) $  cannot be predetermined,
but should be updated adaptively with the training process.
To save computational cost,
this section will present a method to compute $\beta(t)$ based on the sub-training set.

First, notice that even with the same network structure,
the amplitude of the training loss can vary greatly from equation to equation.
It is hard to select a threshold that applies to all equations directly through training losses.
However, we find that for singular perturbation equations,
the gradient of the loss curve is extremely steep around the layer (Figure \ref{DomLoss}).
Therefore, we argue that indirectly determining the threshold $\beta(t)$ by the gradient  of the loss curve
can make the method have a better versatility.

More specifically, let $ X_{sub} $ be a randomly selected subset from the training set,
which is fixed during the training process.
Let $ G $ be a predefined hyperparameter.
After the $t$-th iteration,
for each $ \bm{x}\in X_{sub} $,
if  $ |\nabla_{\bm{x}} r_{phys}^2(\bm{x};\theta)| < G $,
which means that the collocation point is on the outside of the layer region,
then we store its training loss in a memory bank $ M $.
Finally, the maximum loss value in $ M $ is chosen as the threshold $ \beta(t) $.
In this way, $\beta(t)$ can be considered as an upper bound of the training losses of all non-layer samples.
If the loss of a sample exceeds this threshold,
the sample is considered to be close to layer regions and its weight needs to be reduced in the next training iterations.

\begin{rmrk}
Since the training loss usually does not change quickly,
especially in later training periods,
to save computational cost,
we employ an interval update strategy, where the threshold $\beta(t)$ is updated every $K$ iterations.
In our experiments,  $ K $ is set as $ 50 $.
\end{rmrk}

\begin{rmrk}
The proposed approach falls under the category of curriculum learning \cite{Bengio2009,Hac2019},
which mimics human learning and suggests neural networks to prioritize learning easier tasks.
Our algorithm incorporates the properties of singularly perturbed equations and therefore is distinctly different from those existing  curriculum learning algorithms,
which are mainly developed for computer vision \cite{Guo2018,Sara2018} and  natural language processing \cite{Liu2020,Xu2020}.
\end{rmrk}

The pseudo-code of  the proposed approach is summarized in Algorithm 1.
\begin{algorithm}[!htb]
  \caption{Pseudo-code of the curriculum learning for singularly perturbed problems}
  \label{alg}
  \begin{algorithmic}[1]
    \Require
    Training set $X_{train}$,
    subset $X_{sub} \subset X_{train} $,
    predefined constant $ G $,
    balance parameter $\lambda $,
    and update frequency $ K $.
   \State Initialize the iteration step $ t=0 $.
   \For{each training step $ t $}
    \If{ $ t $ is divisible by $K$}
            \State Clean the memory bank $ M $.
            \For{each collocation point $ \bm{x} \in X_{sub} $}
                       \If { $ |\nabla_{\bm{x}} r_{phys}^2(\bm{x};\theta)| < G $}
                                \State  Store the corresponding training loss into $ M $.
                       \EndIf
            \EndFor
            \State Update the threshold $\beta(t) $ by  the maximum loss in the bank $ M $.
    \EndIf
    \State Update the sample weights by \eqref{SampleWeight}.
    \State Update the network parameters based on the loss functions \eqref{bloss}  and \eqref{Mploss}.
    \State t++.
    \EndFor
 \end{algorithmic}
\end{algorithm}

\section{Experiments}
\label{section-exp}

\subsection{Experimental setup}

To evaluate the performance of the proposed method,
six benchmark convection-diffusion-reaction equations,
including one 1-dimensional example, three 2-dimensional examples, and one 3-dimensional example, are considered.
We implement our approach with PyTorch and run the experiments on a Intel Xeon CPU E5-2650 v3 platform with 14GB ROM and a RTX 3060 GPU.
The balance weight $ \lambda $ in the optimization objective \eqref{resloss} is set to 1,
the update frequency $ K=50 $,
and the constant $ G $ is set to 10 in the one-dimensional case and to 50 in the multidimensional cases.
The subset $ X_{sub} $ is 1/5 of the size of the entire training set.

We utilize six fully connected feedforward neural networks to solve different equations, respectively.
All networks employ the Tanh function as the activation unit.
Training process is performed using the Adam optimizer \cite{King2015}.
The specific network structures as well as the training parameters are specified in Table \ref{TNN}.
\begin{table}[!htb]
    \center
    \caption{Structures of neural networks and learning parameters.}
    \label{TNN}
    \scalebox{0.8}{
        \begin{tabular}{c|cccccc}
            \hline
            Equation       &Network depth              &Network width &Optimizer  &Batch size  &Learning rate         &Iterations           \\
            \hline
            5.1               &3                                   &20                   &Adam        &50             &0.001                     &$1.5 \times 10^5$    \\
            \hline
            5.2               &5                                   &20                   &Adam        &200           &0.01                       &$1.5 \times 10^6$    \\
            \hline
            5.3               &3                                   &20                   &Adam        &200           &0.01                       &$1 \times 10^6$      \\
            \hline
            5.4               &3                                   &20                   &Adam        &200           &0.01                       &$1 \times 10^6$      \\
            \hline
            5.5               &3                                   &20                   &Adam        &200           &0.005                     &$1.5 \times 10^6$    \\
            \hline
            5.6               &5                                   &20                   &Adam        &500           &0.01                       &$1 \times 10^6$      \\
            \hline
    \end{tabular}}
\end{table}

For the one dimensional equation, we employ a uniform sampling to construct the training set.
For the multi-dimensional problems,
to ensure that there are a number of training points belonging to the layer regions,
we adopt a non-uniform sampling.
Specifically, we first uniformly sample half of the training points,
and then add more samples around the points whose feedforward losses exceed the threshold $ \beta(t)$,
until the training set reaches the predefined size.
The size of the training set for each equation is listed in Table \ref{Tsample}.

\begin{table}[htb]
    \center
    \caption{Number of training points for different equations.}
    \label{Tsample}
    \scalebox{0.8}{
        \begin{tabular}{c|cc}
            \hline
            Equation           &Interior samples                &Boundary samples             \\
            \hline
            5.1                   &$2.5 \times 10^3$             &2                            \\
            \hline
            5.2--5.5           &$2 \times 10^4$                &$4 \times 10^2$               \\
            \hline
            5.6                  &$3 \times 10^5$                &$6 \times 10^4$              \\
            \hline
    \end{tabular}}
\end{table}

If the exact solution is known,
we quantify the performance of the prediction by using the Normalized Root-Mean-Squared Error (NRMSE):
\begin{align*}
  \text{NRMSE}=\frac{\sqrt{\sum_{i=1}^n |u_{\theta}(x_i)-u(x_i)|^2}}{\sqrt{\sum_{i=1}^n |u(x_i)|^2}},
\end{align*}
where $ u_\theta(x)  $ and $ u(x) $ represent the predicted and the exact solution, respectively,
and $ n $ denotes the number of uniformly sampled test points,
which is set to 1000 for one-dimensional equation, and 5000 for multi-dimensional equations.

\subsection{1d convection-diffusion equation}

Consider the following two point problem:
\begin{align}
\label{1de}
\begin{split}
   -\epsilon u_{xx}+(x-2) u_x &=f(x), \quad x\in (0,1),
   \\[5pt]
   u(0)=u(1)&=0,
\end{split}
\end{align}
where the source term $ f(x) $ is chosen such that the exact solution $u(x)=\cos(\pi x/2)(1-\exp(-2x/\epsilon)) $.
The solution of \eqref{1de} is characterized by a boundary layer at $ x=0 $.

\begin{figure}[!hbt]
\centering
\subfloat[Training loss curves]{\includegraphics[width=0.33\textwidth]{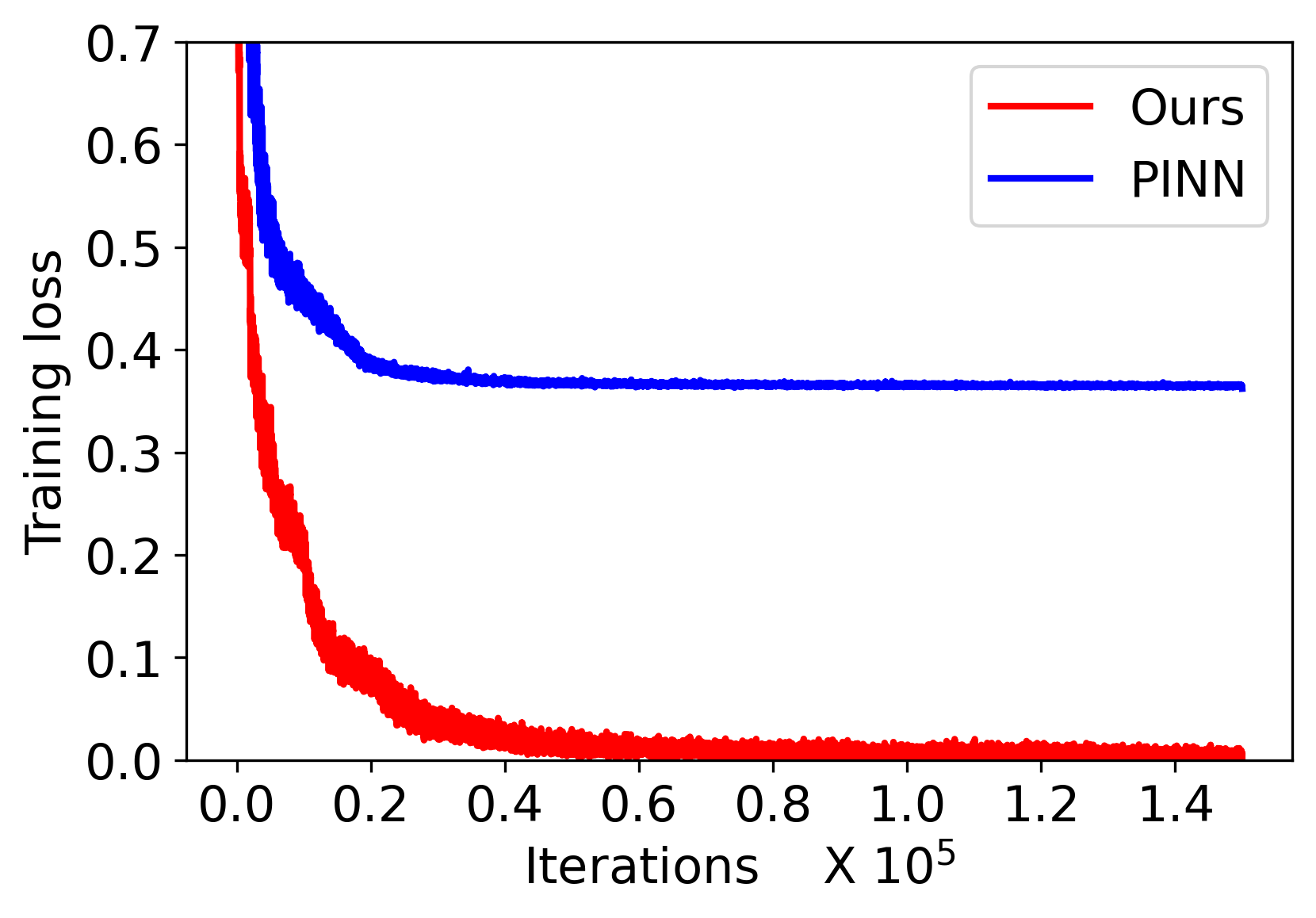}}\hfill
\subfloat[Predictions ($\epsilon = 1e-9$)]{\includegraphics[width=0.33\textwidth]{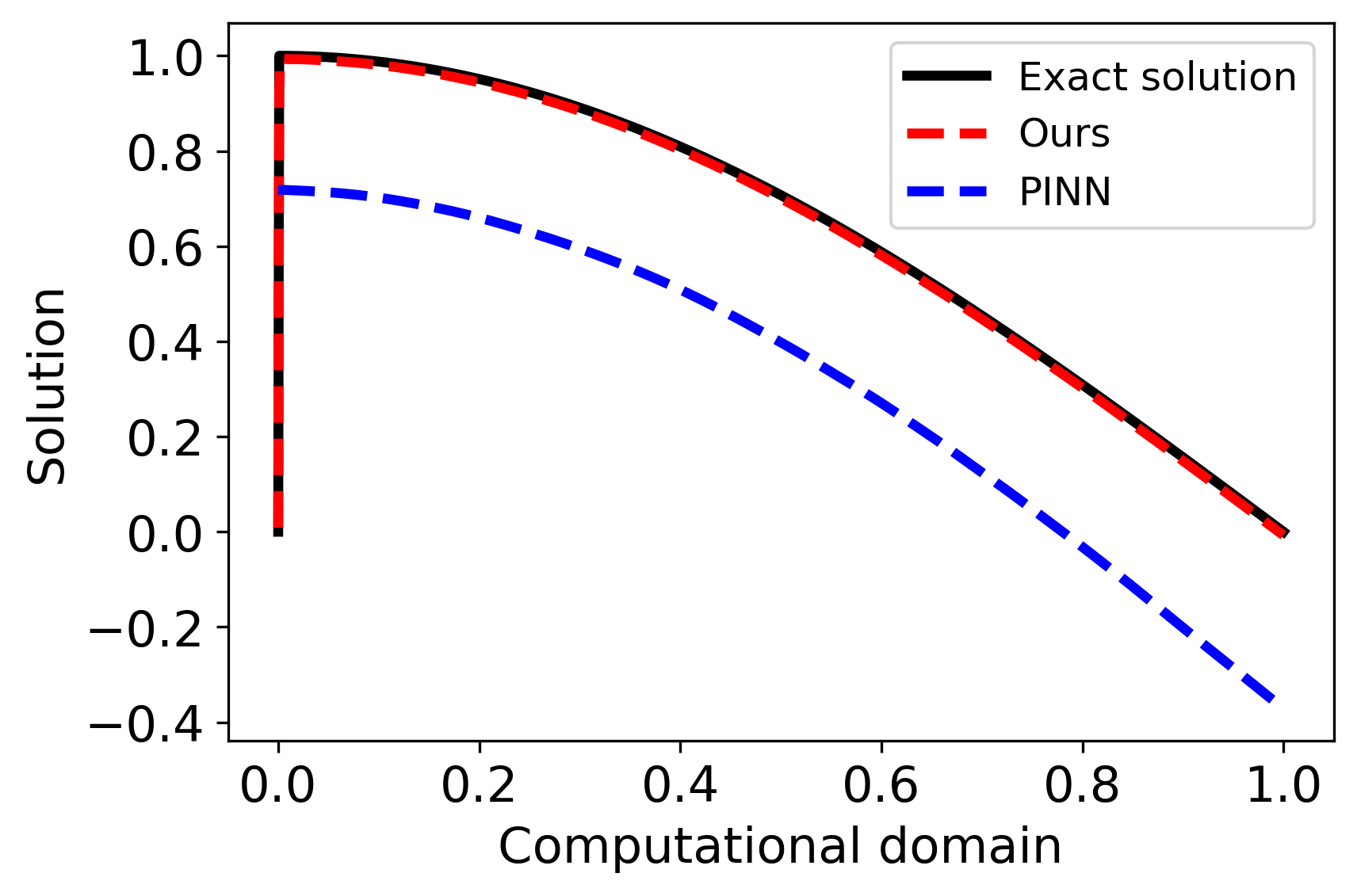}}\hfill
\subfloat[Absolute errors]{\includegraphics[width=0.33\textwidth]{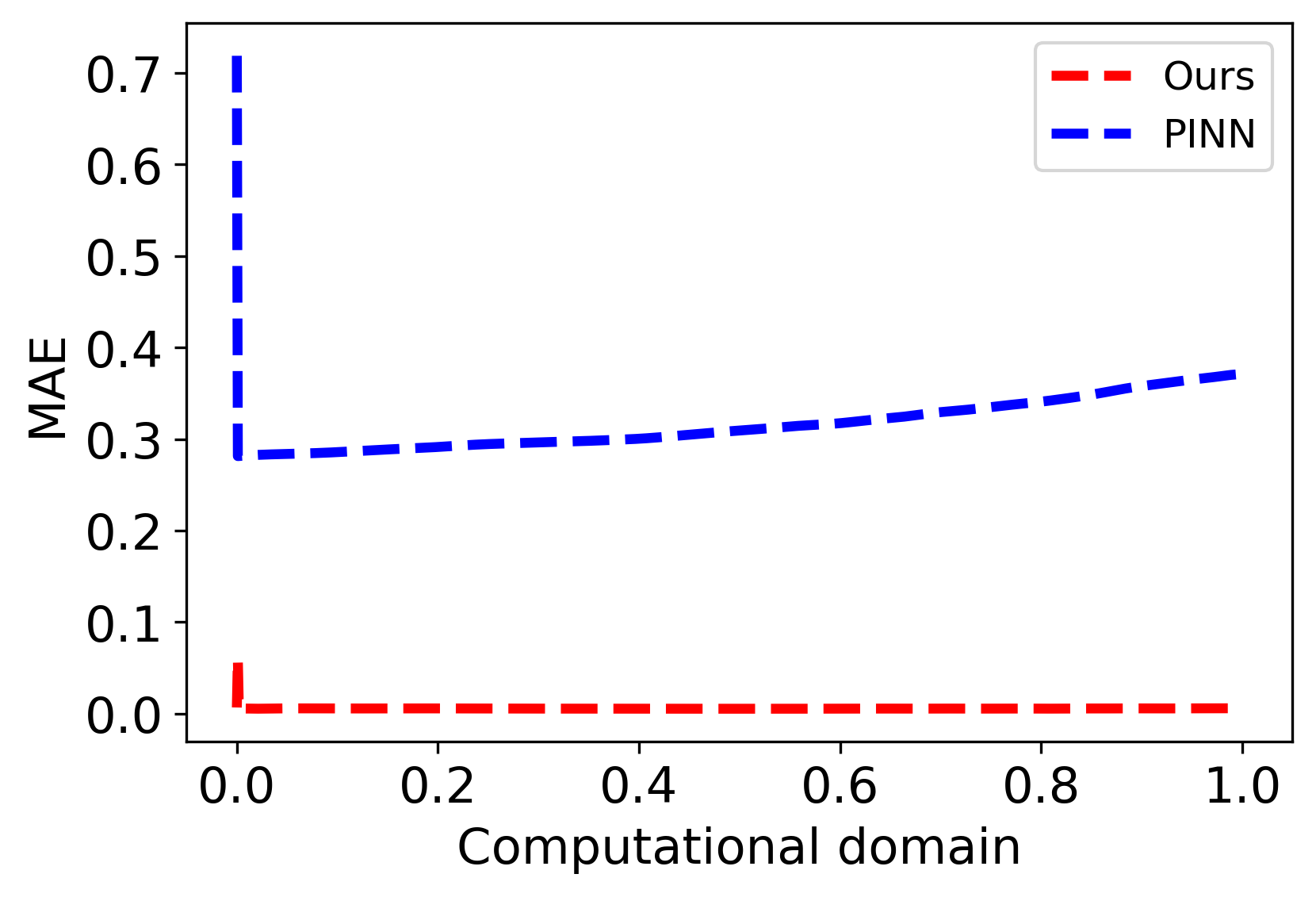}}\hfill
\caption{Comparison between our approach and PINN for one dimensional equation \eqref{1de} with $\epsilon=1e-9$.}
\label{figure1de}
\end{figure}

We first plot the training loss curves of our approach and the conventional PINN in the case of $ \epsilon=1e-9 $.
As demonstrated in Figure \ref{figure1de} (a),
our method descends much faster than PINN in the early stage of training,
and the corresponding training loss approaches 0 after about 4000 iterations.
In contrast, PINN fails to converge even after a long period of iterations.
It can be further observed from Figure \ref{figure1de} (b) and Figure \ref{figure1de} (c) that
the prediction of our approach captures the boundary layer well  and fits the exact solution much better over the entire computational domain,
while the result of PINN differs significantly.

Further, we compare the normalized root-mean-squared errors of the two methods with more diffusion coefficients.
It is obvious from Table \ref{table1de} that for non-singularly perturbed case ($\epsilon=1$),
both methods can produce satisfactory results of the same order of accuracy.
However, for singularly-perturbed cases,
the errors of our method is 3 orders of magnitude lower than those of PINN.

\begin{table}[!htb]
    \center
    \caption{Normalized root-mean-squared errors between the predicted and exact solutions of \eqref{1de} under various diffusion coefficients.}
    \label{table1de}
    \scalebox{1.0}{
        \begin{tabular}{c|cc}
            \hline
            Diffusion coefficient          &Ours                      &  PINN                 \\
                \hline
            $\epsilon= 1 $                 &$1.81 \times 10^{-4}$     &$1.83 \times 10^{-4}$  \\
            \hline
            $\epsilon= 1e-3 $              &$1.41 \times 10^{-4}$     &$4.45 \times 10^{-1}$  \\
            \hline
            $\epsilon= 1e-6 $              &$1.44 \times 10^{-4}$     &$4.54 \times 10^{-1}$  \\
            \hline
            $\epsilon= 1e-9 $              &$1.47 \times 10^{-4}$     &$4.47 \times 10^{-1}$  \\
            \hline
    \end{tabular}}
\end{table}




\subsection{2d convection-diffusion-reaction equation with boundary layers}

Consider the following two-dimensional problem \cite{Zhang2016}:
\begin{align}
\label{2dbl}
\begin{split}
     -\epsilon \Delta u +(3-x_1-x_2)u_{ x_1}+1.5 u&=f,\quad  \bm{x}\in \Omega=(0,1)^2,
     \\[5pt]
     u&=0,\quad  \bm{x} \in \partial \Omega,
\end{split}
\end{align}
where $ f(x) $ is chosen such that the exact solution
\begin{align*}
   u=\bigg(\sin\frac{\pi x_1}{2}-\frac{e^{-(1-x_1)/\epsilon}-e^{-1/\epsilon}}{1-e^{-1/\epsilon}}\bigg)\frac{(1-e^{-x_2/\sqrt{\epsilon}})(1-e^{-(1-x_2)/\sqrt{\epsilon}})}{1-e^{-1/\sqrt{\epsilon}}}.
\end{align*}
The solution of \eqref{2dbl} is characterized by the presence of three boundary layers,
one at $x_1=1 $, and two at $ x_2=0 $ and $ x_2=1 $.

\begin{figure}[!hbt]
\centering
\subfloat[Exact solution]{\label{fig:mdleft}{\includegraphics[width=0.33\textwidth]{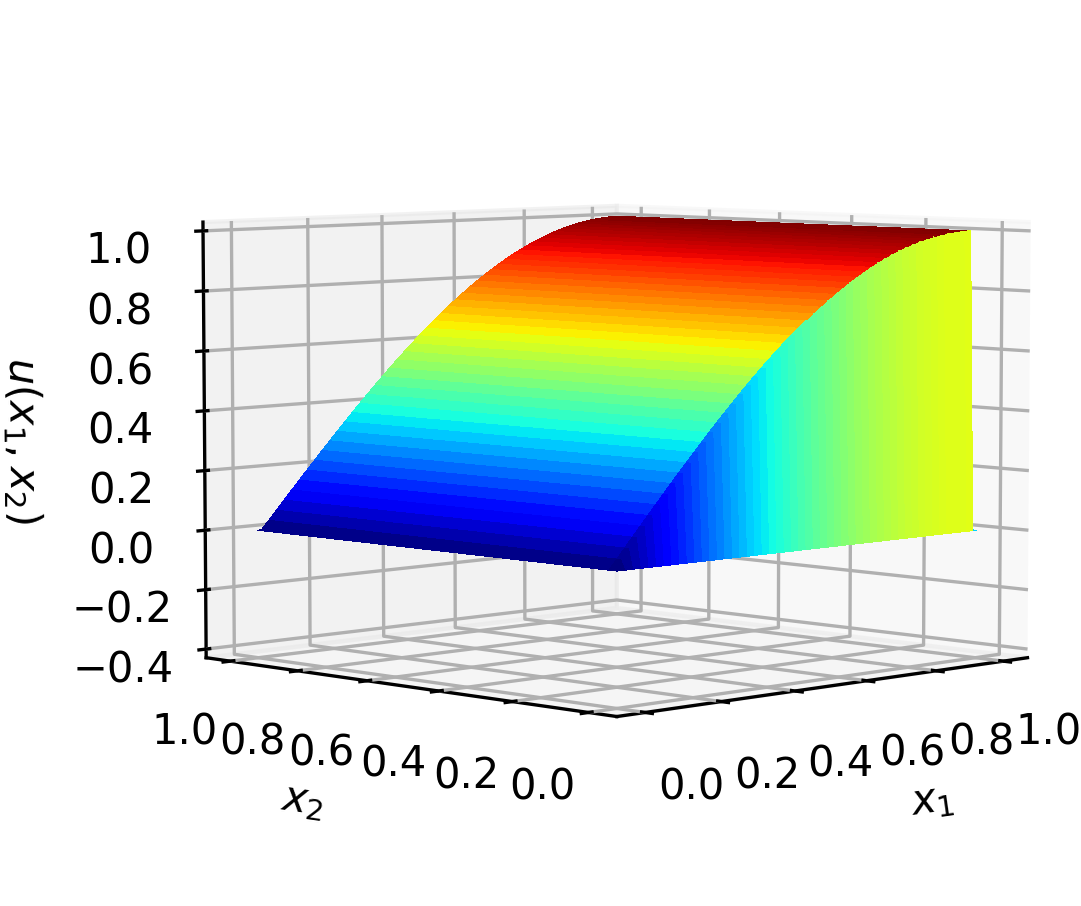}}}\hfill
\subfloat[Our prediction]{\label{fig:mdleft}{\includegraphics[width=0.33\textwidth]{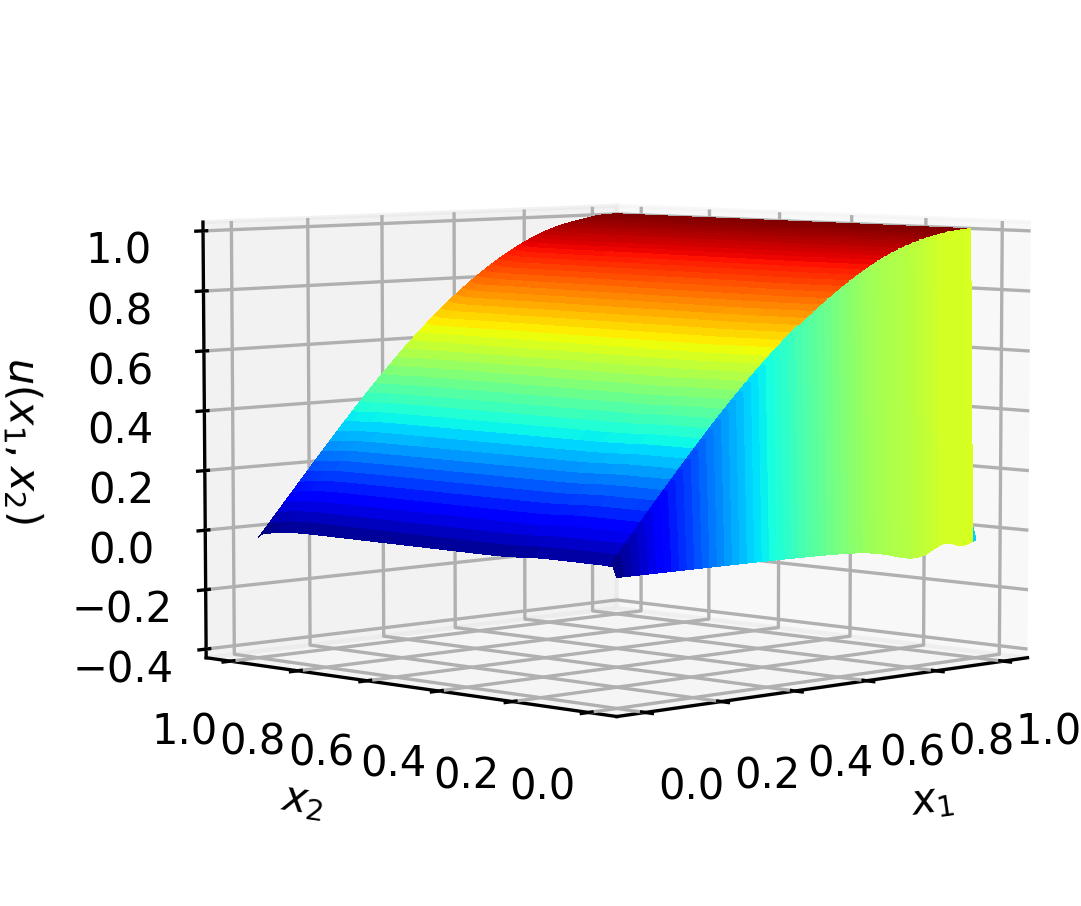}}}\hfill
\subfloat[PINN prediction]{\label{fig:mdleft}{\includegraphics[width=0.33\textwidth]{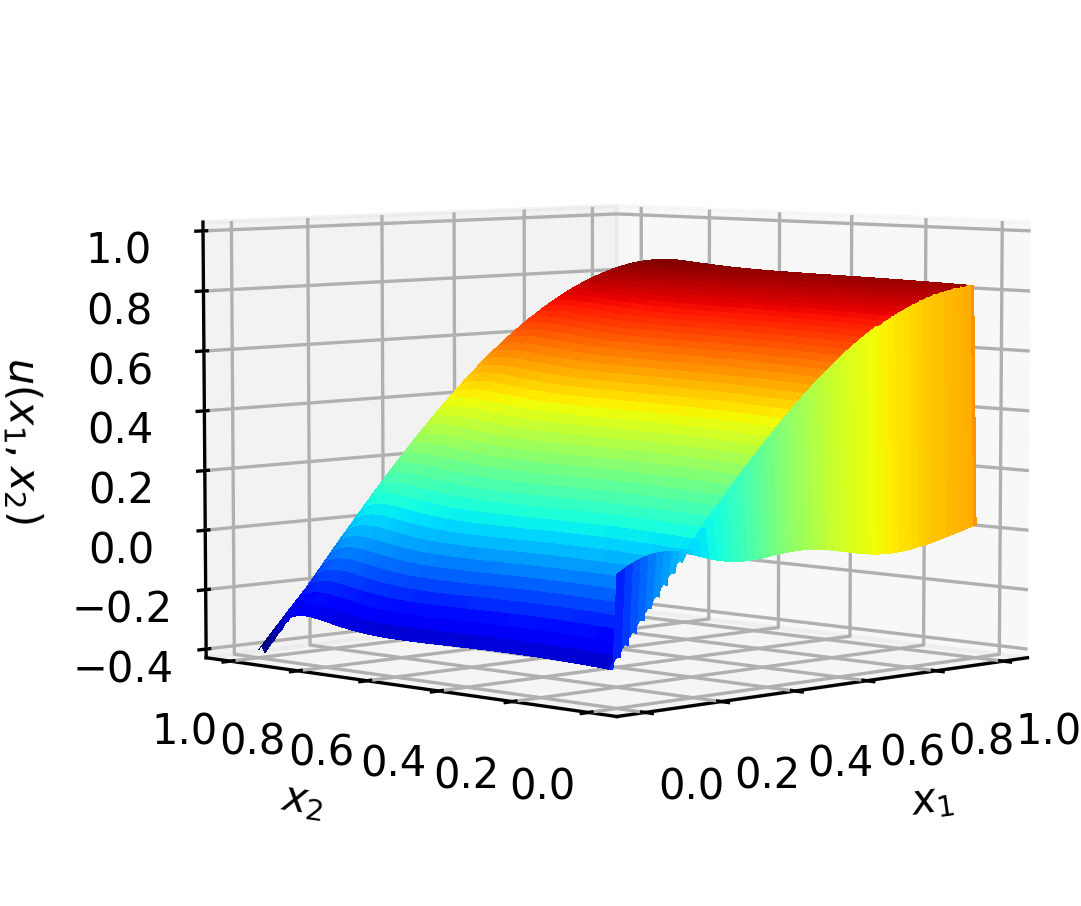}}}\hfill
\subfloat[Training loss curves]{\label{fig:mdleft}{\includegraphics[width=0.33\textwidth]{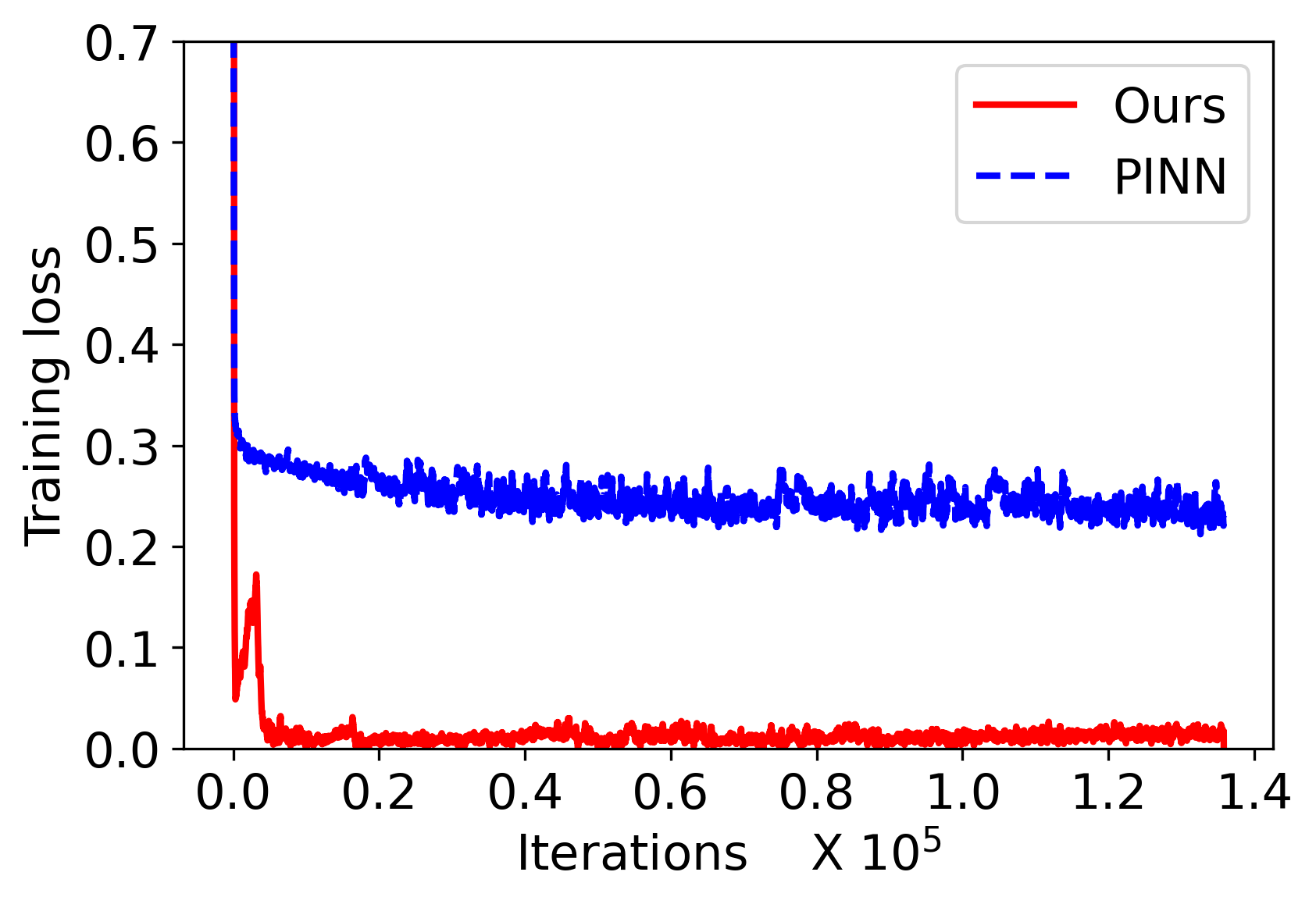}}}\hfill
\subfloat[Absolute errors of our approach]{\label{fig:mdleft}{\includegraphics[width=0.33\textwidth]{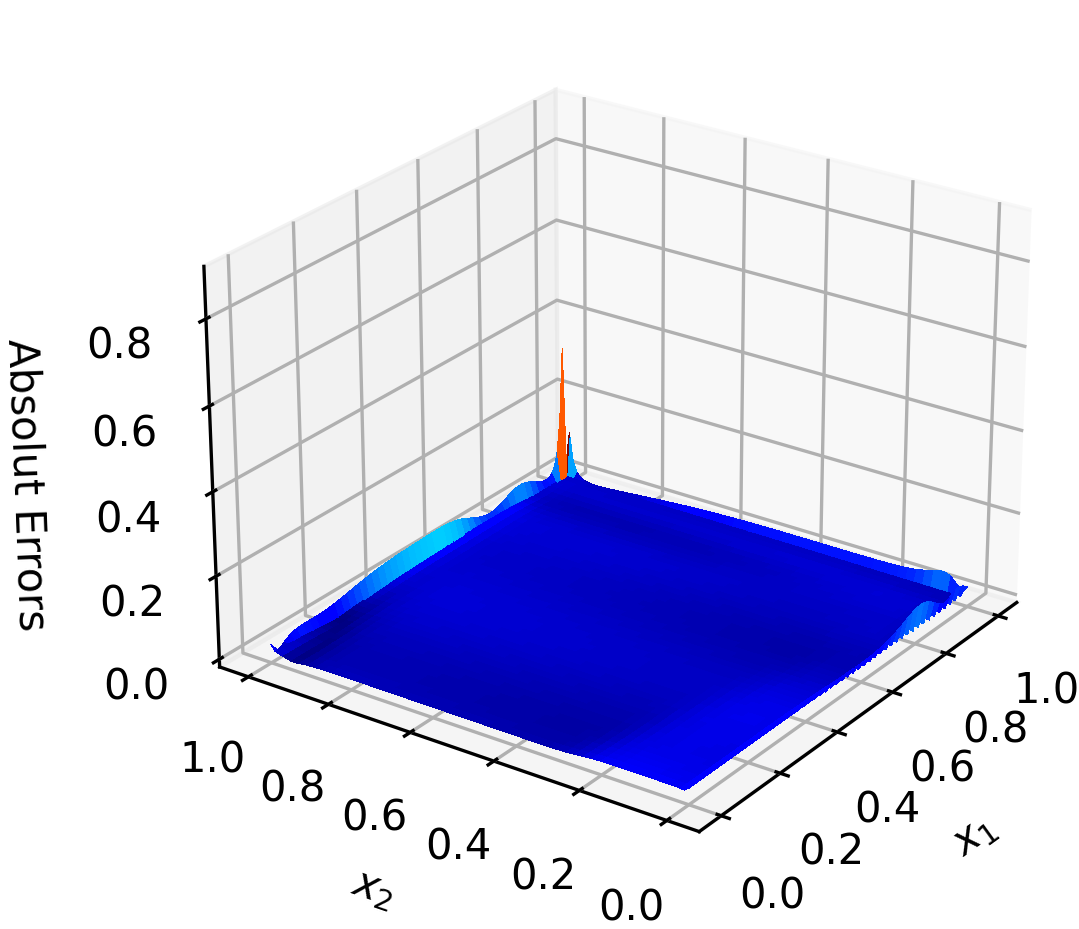}}}\hfill
\subfloat[Absolute errors of PINN]{\label{fig:mdleft}{\includegraphics[width=0.33\textwidth]{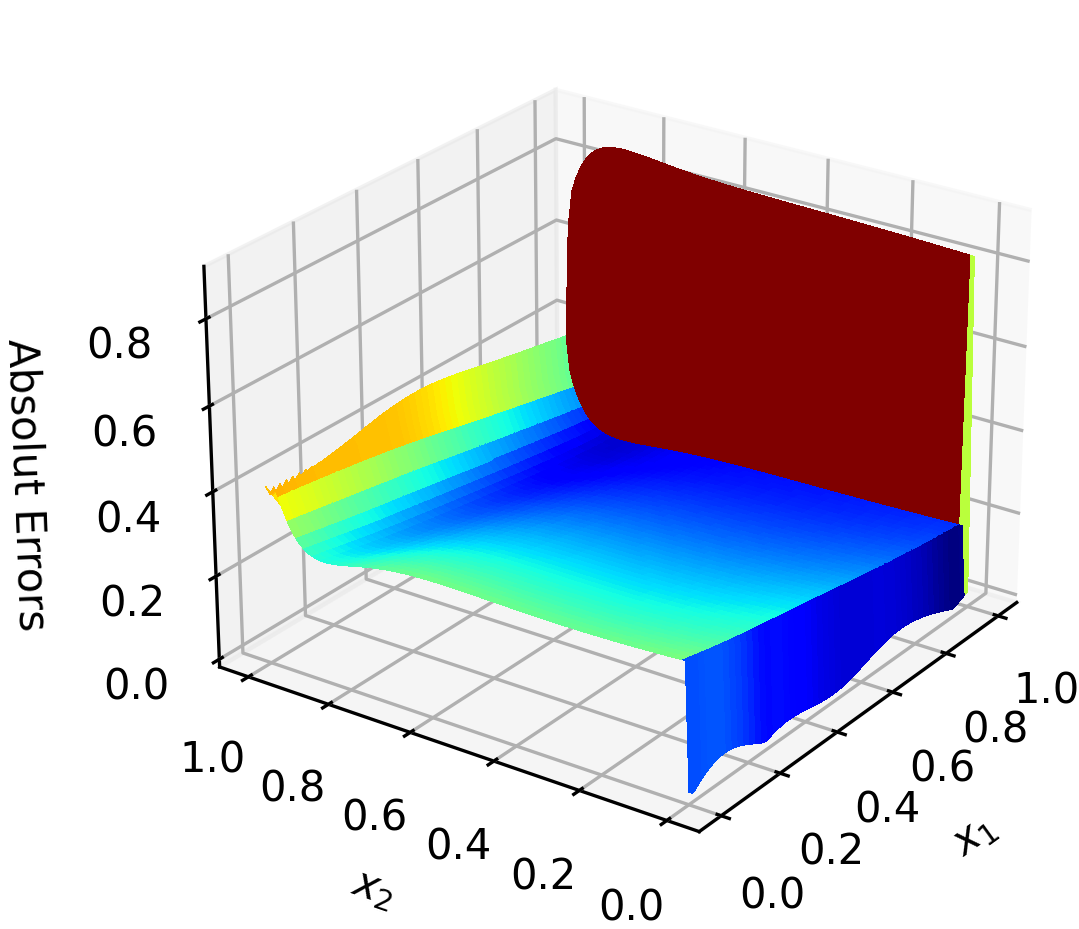}}}\hfill
\caption{Comparison between our approach and PINN for equation \eqref{2dbl} with $\epsilon = 1e-9$.}
\label{figure2dbl}
\end{figure}

It can been observed from Figure \ref{figure2dbl} and Table \ref{table2dbl} that
our method still performs well in capturing the behavior of the layers.
The predictions only have a little oscillation at the boundary layer location.
In contrast, the common PINN shows a considerable deviation from the truth.

\begin{table}[!htb]
    \center
    \caption{Normalized root-mean-squared errors between the predicted and exact solutions of \eqref{2dbl} under various diffusion coefficients.}
    \label{table2dbl}
    \scalebox{1.0}{
        \begin{tabular}{c|cc}
            \hline
            Diffusion coefficient         &Ours                     &  PINN             \\
            \hline
            $\epsilon= 1e-3 $             &$4.67 \times 10^{-4}$    &$3.37 \times 10^{-2}$  \\
            \hline
            $\epsilon= 1e-6 $             &$5.38 \times 10^{-4}$    &$5.31 \times 10^{-1}$  \\
            \hline
            $\epsilon= 1e-9 $             &$5.35 \times 10^{-4}$    &$5.30 \times 10^{-1}$  \\
            \hline
    \end{tabular}}
\end{table}

\subsection{2d convection-diffusion equation with interior layers}
This section is devoted to assessing the performance of the proposed approach  in the presence of interior layers.
To this end, consider \cite{Ayuso2009}
\begin{align}
\label{2dil}
\begin{split}
     -\epsilon \Delta u +\bm{b} \cdot \nabla u   &=0,\quad  \bm{x}\in \Omega=(0,1)^2,\\
       u&=\left\{\begin{array}{ll}
                            1, & \text{if}\; x_2=0, \\
                            1, & \text{if} \;  x_1=0, x_2 \leq 1/5,\\
                            0,&  \text{elsewhere on} \; \partial \Omega,
                      \end{array}\right.
\end{split}
\end{align}
where the convection coefficient $ \bm{b}=(1/2,\sqrt{3}/2)^T$.
The latent solution of equation \ref{2dil} presents both internal and external boundary layers.
For most traditional numerical methods, non-physical  oscillations are often observed near the interior layer
caused by the joints of the conflicting discontinuous boundary conditions.

\begin{figure}[hbt]
\centering
\subfloat[Our prediction ($\epsilon = 10^{-3}$)]{\label{fig:mdleft}{\includegraphics[width=0.3\textwidth]{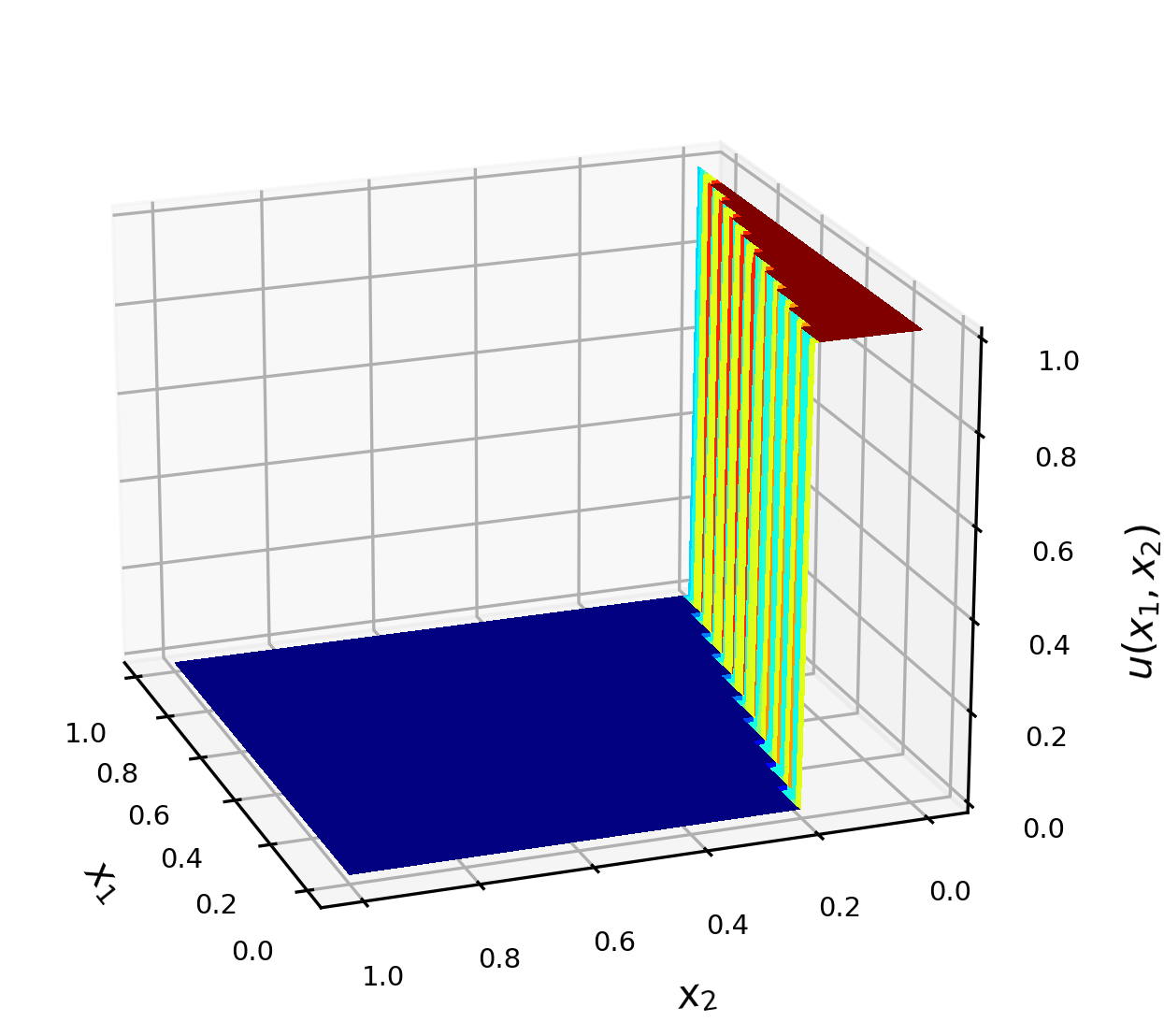}}}\hfill
\subfloat[Our prediction ($\epsilon = 10^{-6}$)]{\label{fig:mdleft}{\includegraphics[width=0.3\textwidth]{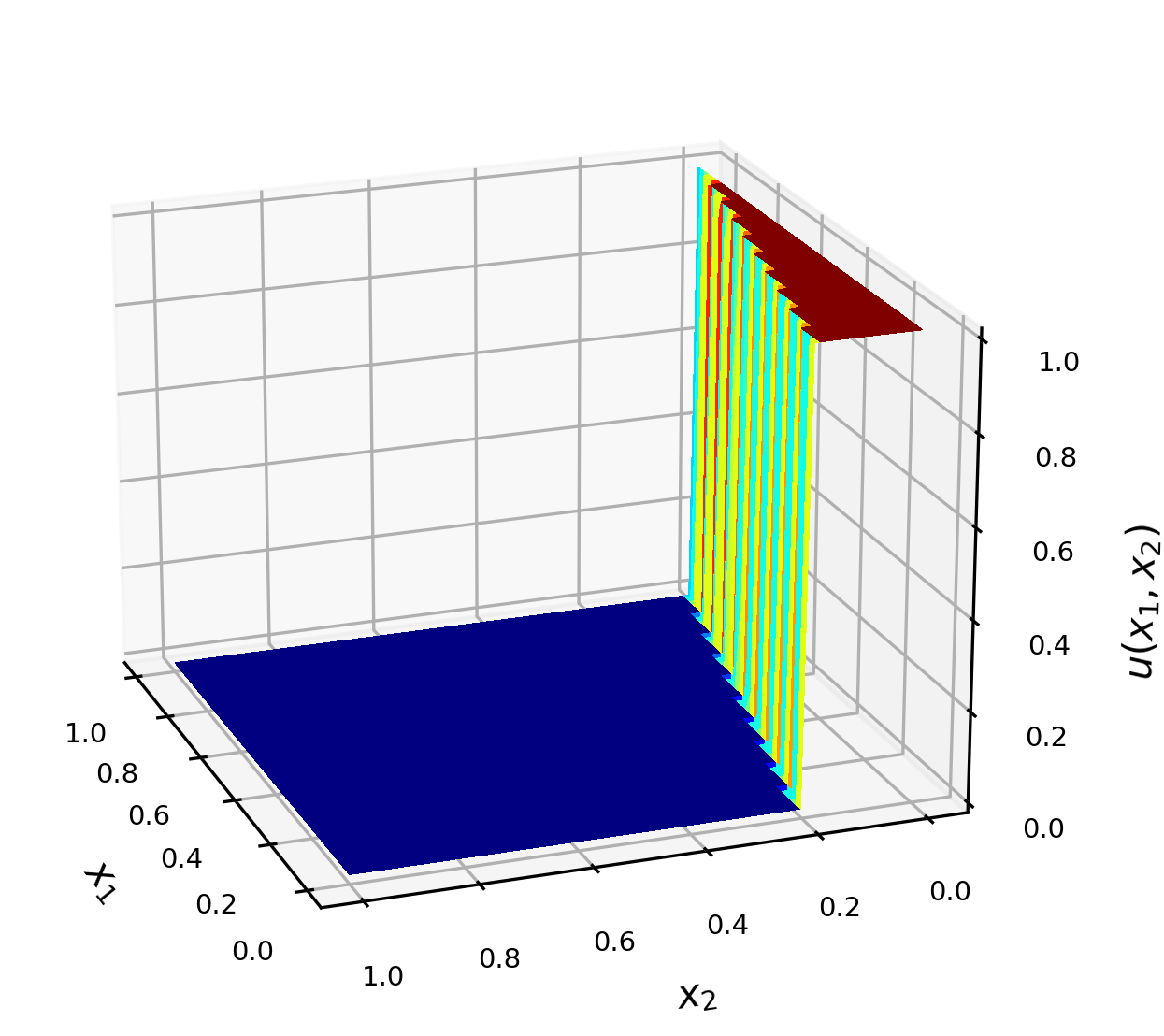}}}\hfill
\subfloat[Our prediction ($\epsilon = 10^{-9}$)]{\label{fig:mdleft}{\includegraphics[width=0.3\textwidth]{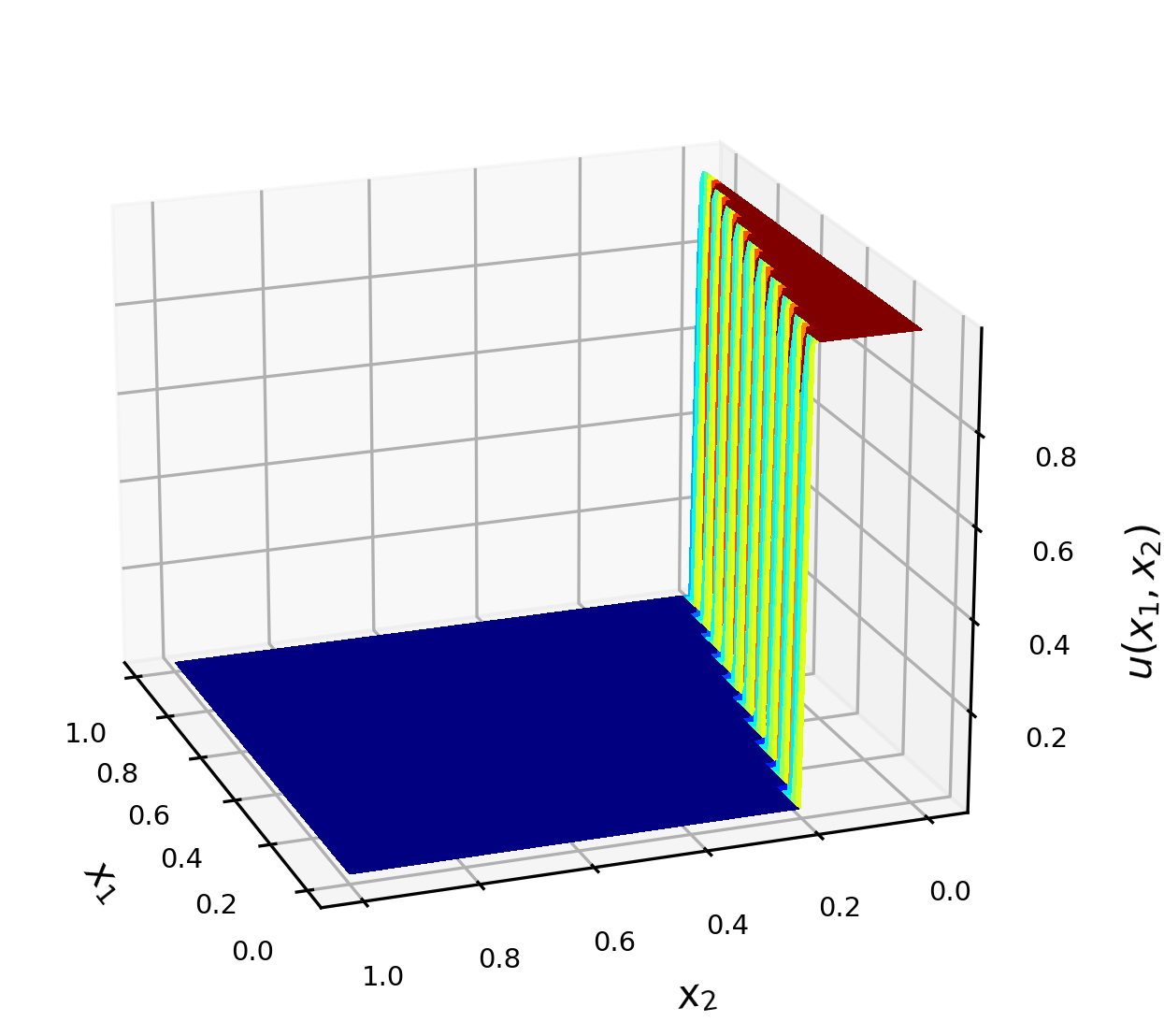}}}\hfill
\subfloat[PINN prediction ($\epsilon = 10^{-3}$)]{\label{fig:mdleft}{\includegraphics[width=0.3\textwidth]{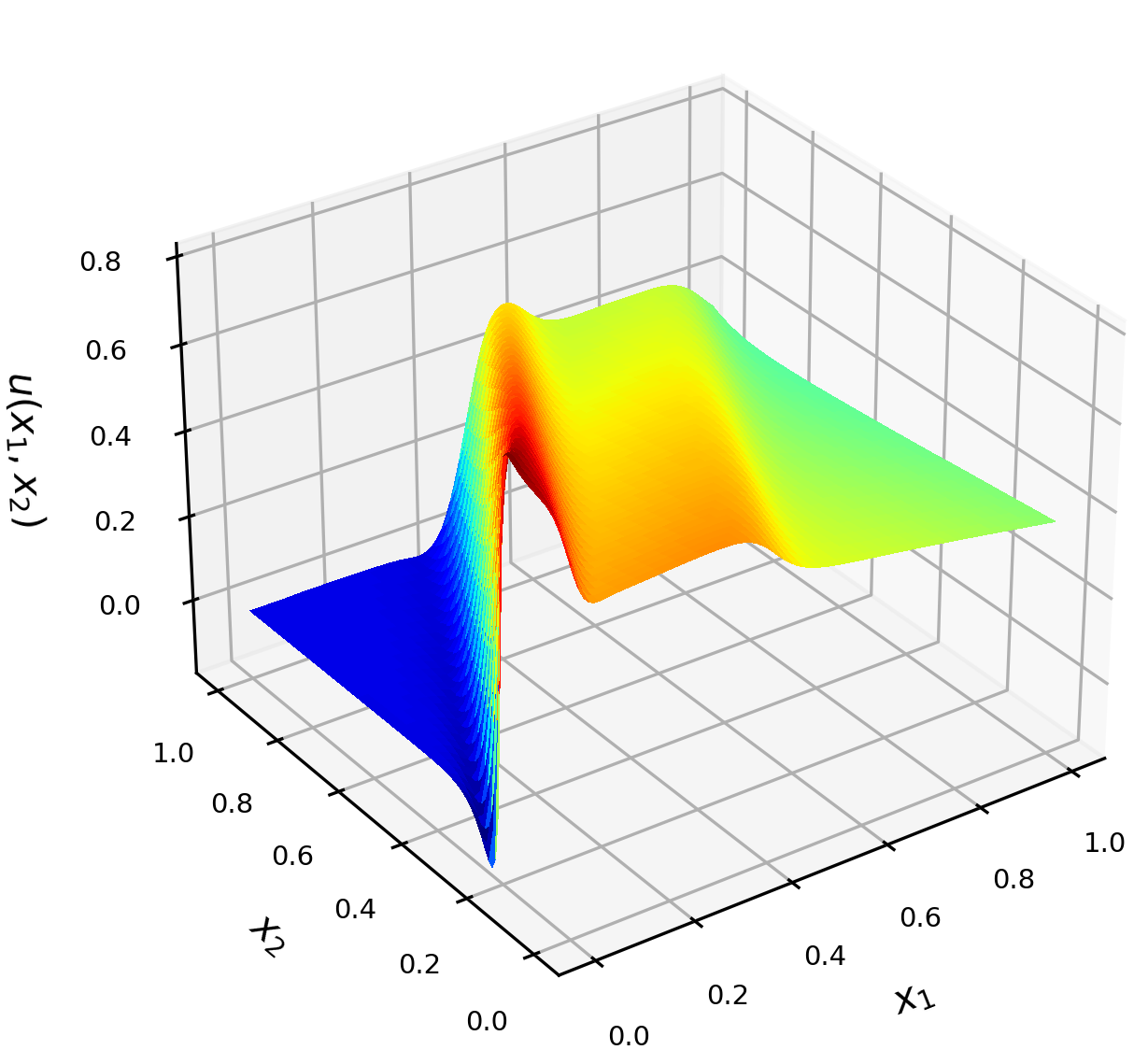}}}\hfill
\subfloat[PINN prediction ($\epsilon = 10^{-6}$)]{\label{fig:mdleft}{\includegraphics[width=0.3\textwidth]{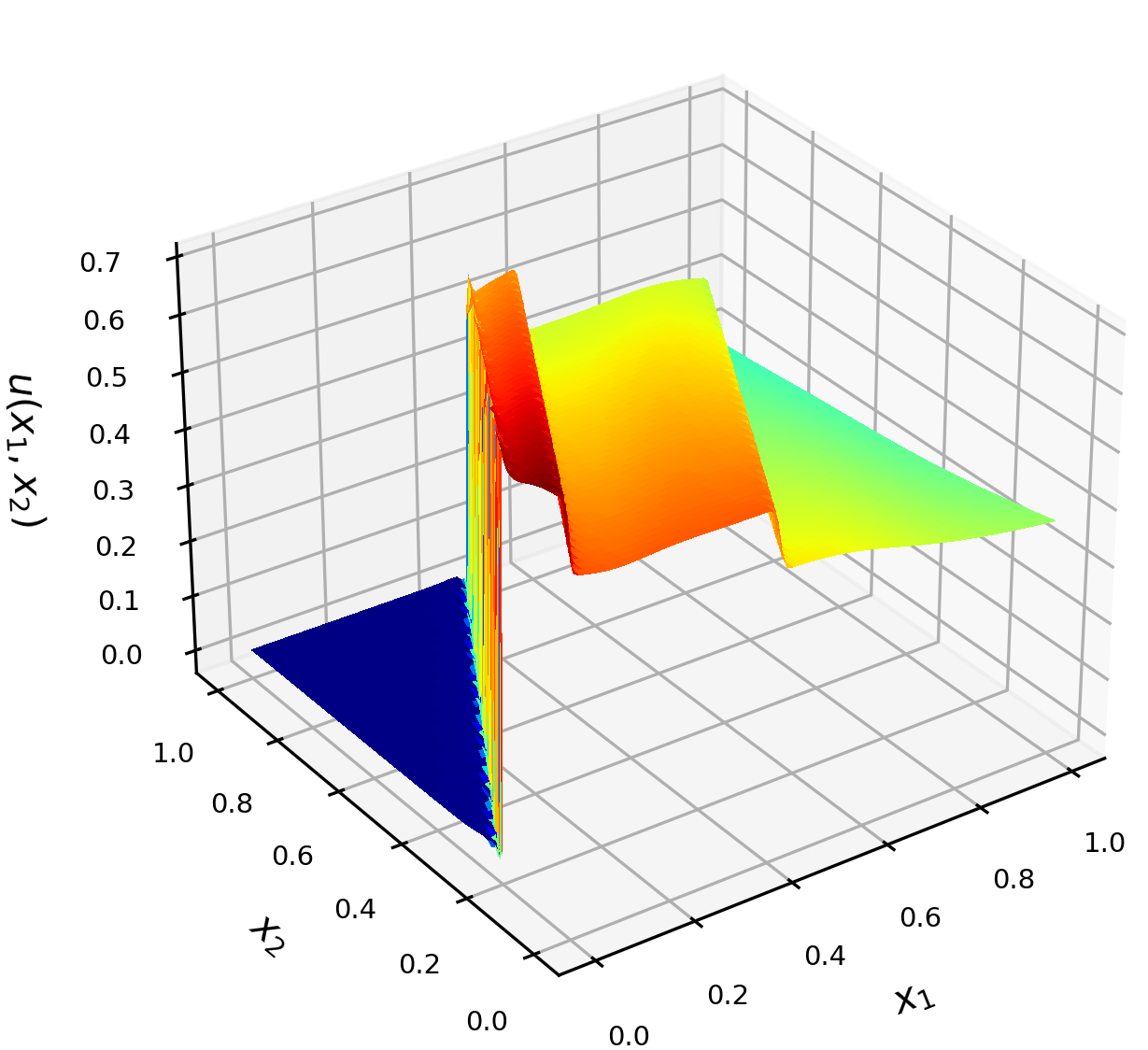}}}\hfill
\subfloat[PINN prediction ($\epsilon = 10^{-9}$)]{\label{fig:mdleft}{\includegraphics[width=0.3\textwidth]{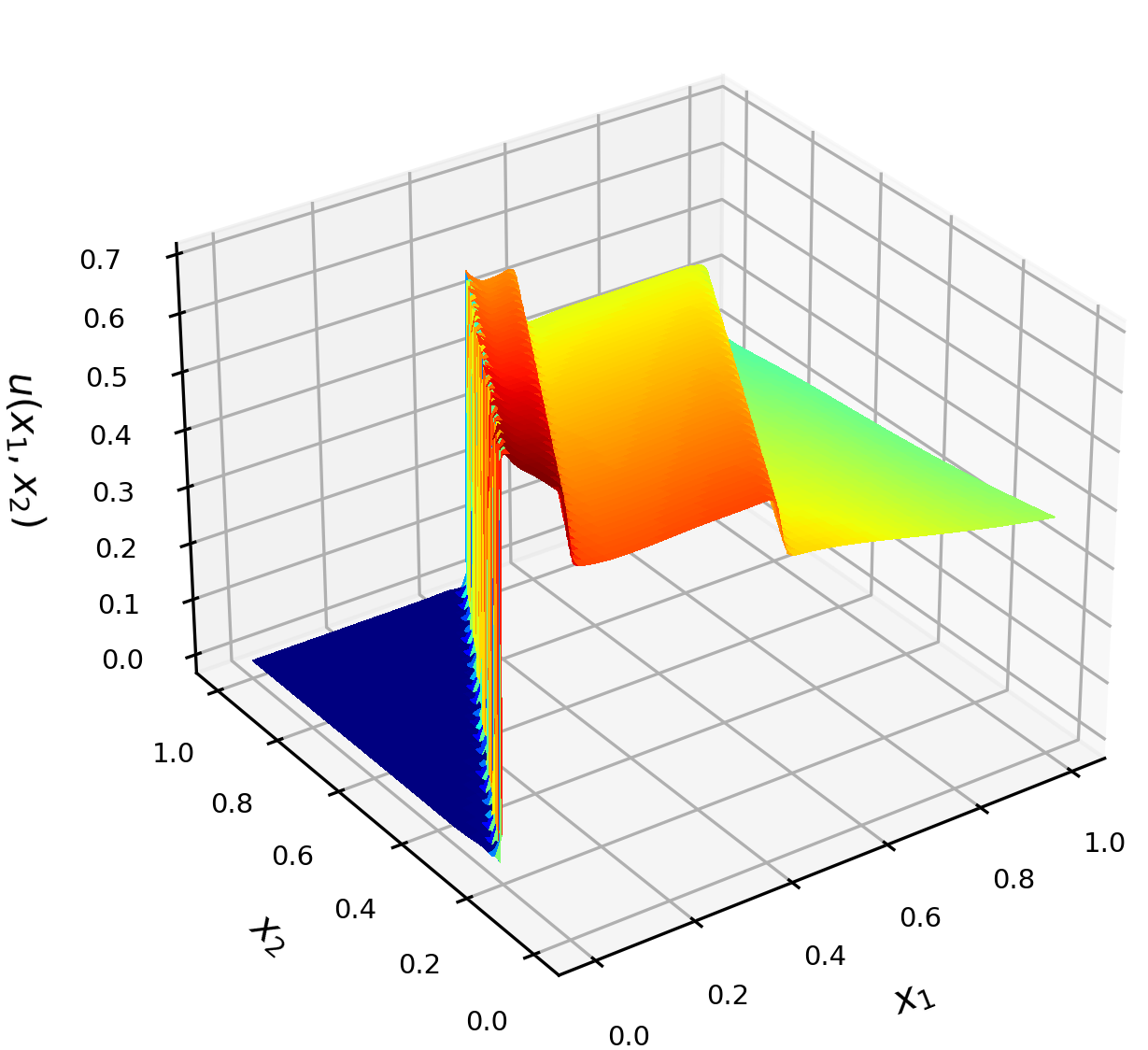}}}\hfill
\caption{Comparison between our approach and PINN  for equation \eqref{2dil} under various diffusion coefficients.}
\label{figure2dil}
\end{figure}

As can be seen from Figure \eqref{figure2dil},
the internal layers are sharply captured by our approach with almost no overshooting/undershooting.
In contrast, common PINN performs very poorly and its predictions are highly oscillatory.
Moreover, in this example, our method is stable with respect to various $\epsilon$.
When $\epsilon$ changes from $1e-3$ to $1e-9$,
 there is no obvious oscillation appearing in the prediction results.

subsection{Rotational flow}
Consider the following rotational flow problem \cite{Hughes2006}:
\begin{align}
\label{2drf}
     -\epsilon \Delta u + \nabla \cdot(\bm{b}u) &=0,\quad  \bm{x}\in \Omega=(0,1)^2,
\end{align}
where the convection coefficient $ \bm{b}=(1/2-x_2,x_1-1/2)^T$,
and the solution is prescribed along the slit $1/2 \times [0, 1/2] $ as follows
\begin{align*}
       u(1/2,x_2)&=\sin^2(2\pi x_2),\quad  x_2 \in [0,1/2].
\end{align*}
The above equation describes the convection of a single component in a rotating flow field,
where the axis of rotation passes through the center of the square domain.

\begin{figure}[!hbt]
\centering
\subfloat[Our prediction ($\epsilon = 10^{-3}$)]{\label{fig:mdleft}{\includegraphics[width=0.33\textwidth]{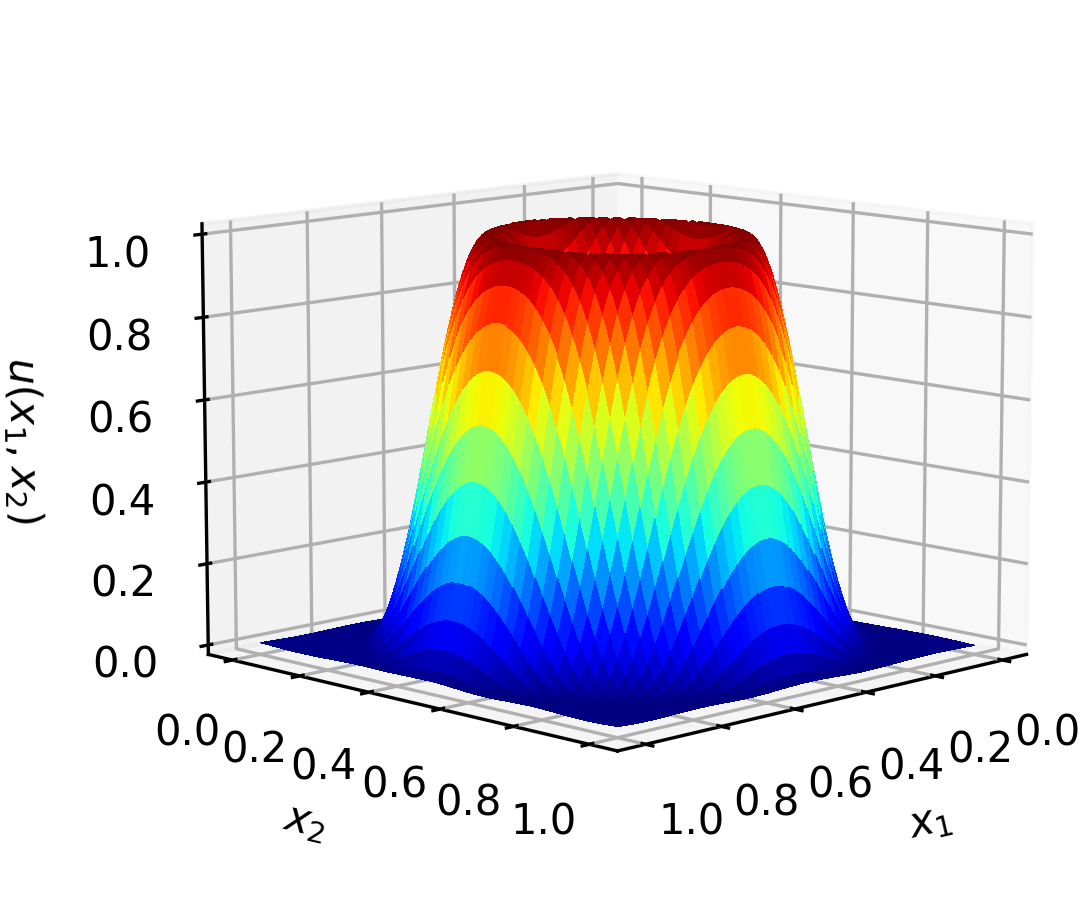}}}\hfill
\subfloat[Our prediction ($\epsilon = 10^{-6}$)]{\label{fig:mdleft}{\includegraphics[width=0.33\textwidth]{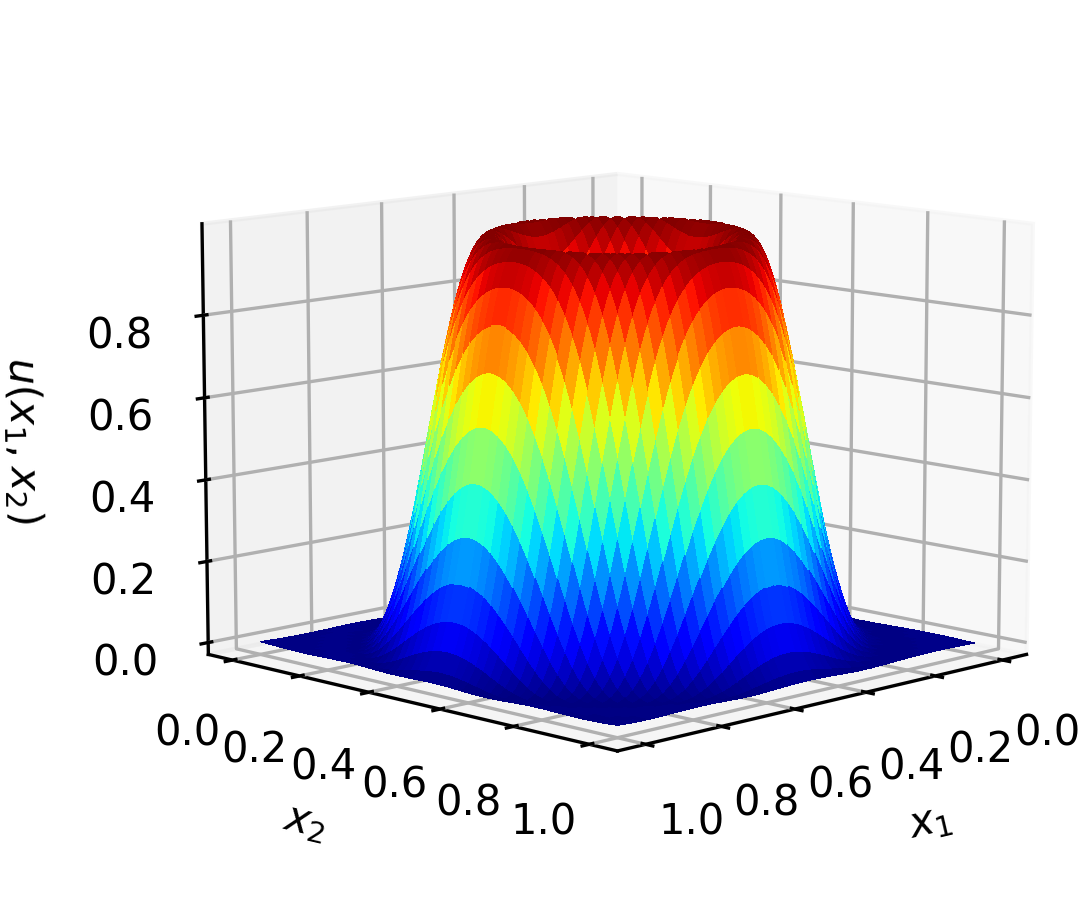}}}\hfill
\subfloat[Our prediction ($\epsilon = 10^{-9}$)]{\label{fig:mdleft}{\includegraphics[width=0.33\textwidth]{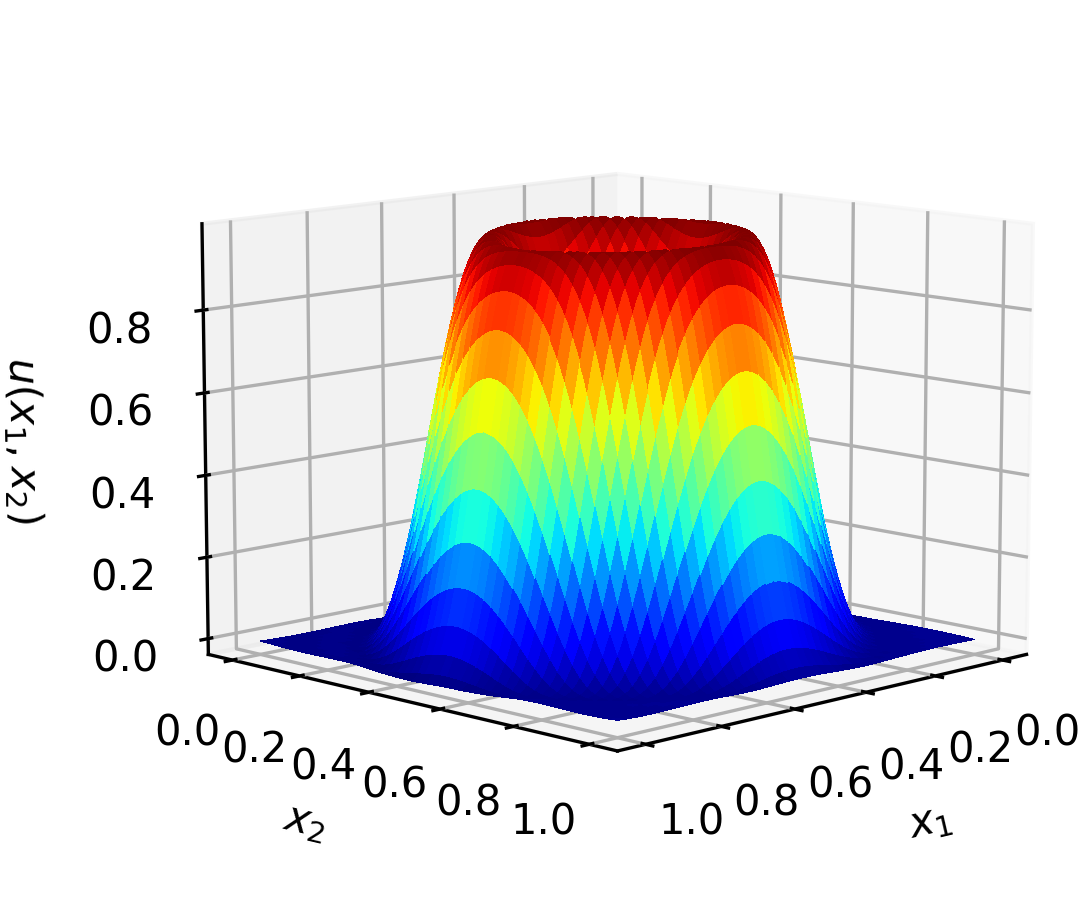}}}\hfill
\subfloat[PINN prediction ($\epsilon = 10^{-3}$)]{\label{fig:mdleft}{\includegraphics[width=0.33\textwidth]{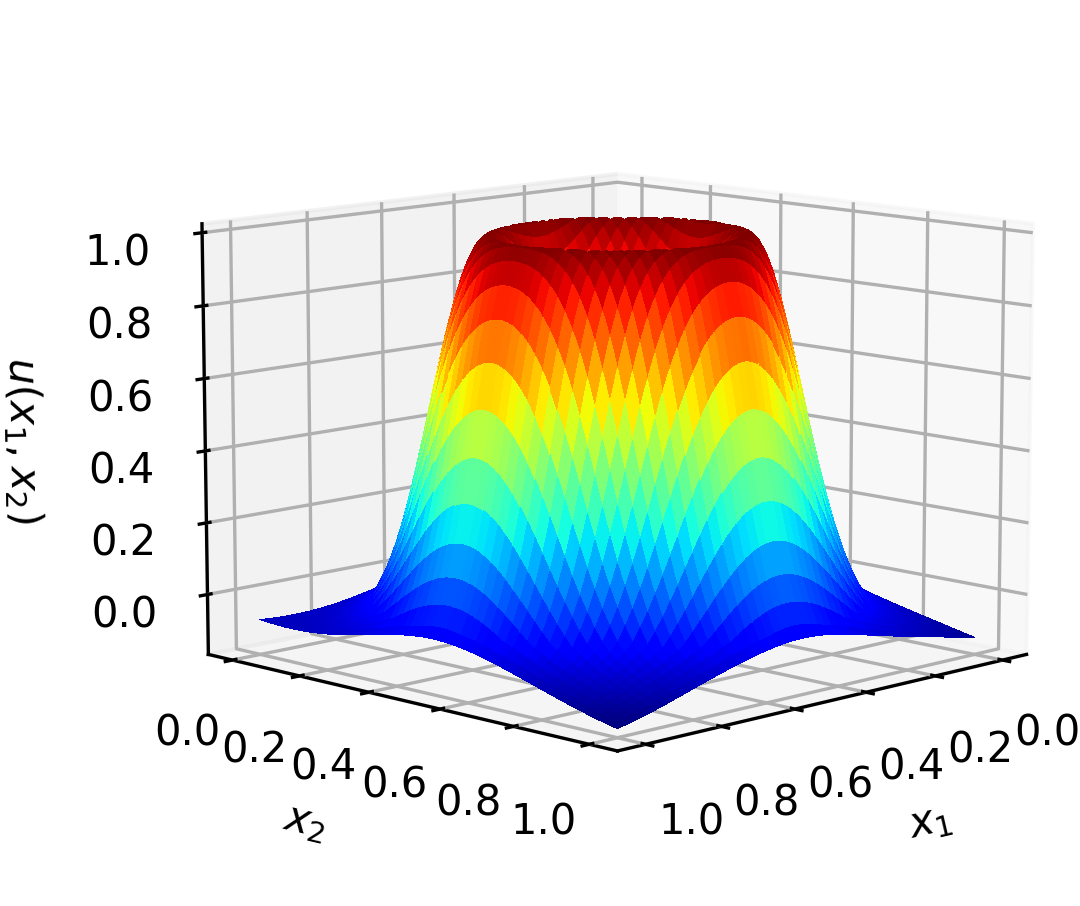}}}\hfill
\subfloat[PINN prediction($\epsilon = 10^{-6}$)]{\label{fig:mdleft}{\includegraphics[width=0.33\textwidth]{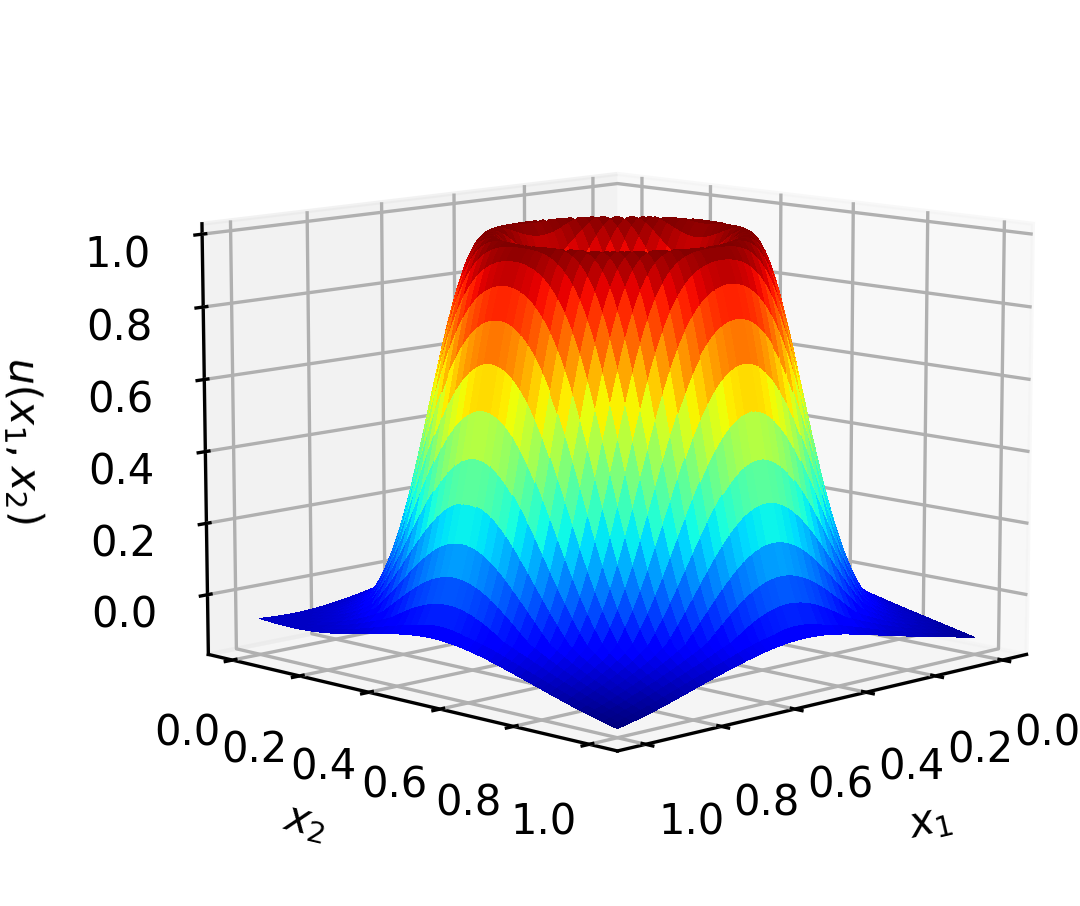}}}\hfill
\subfloat[PINN prediction($\epsilon = 10^{-9}$)]{\label{fig:mdleft}{\includegraphics[width=0.33\textwidth]{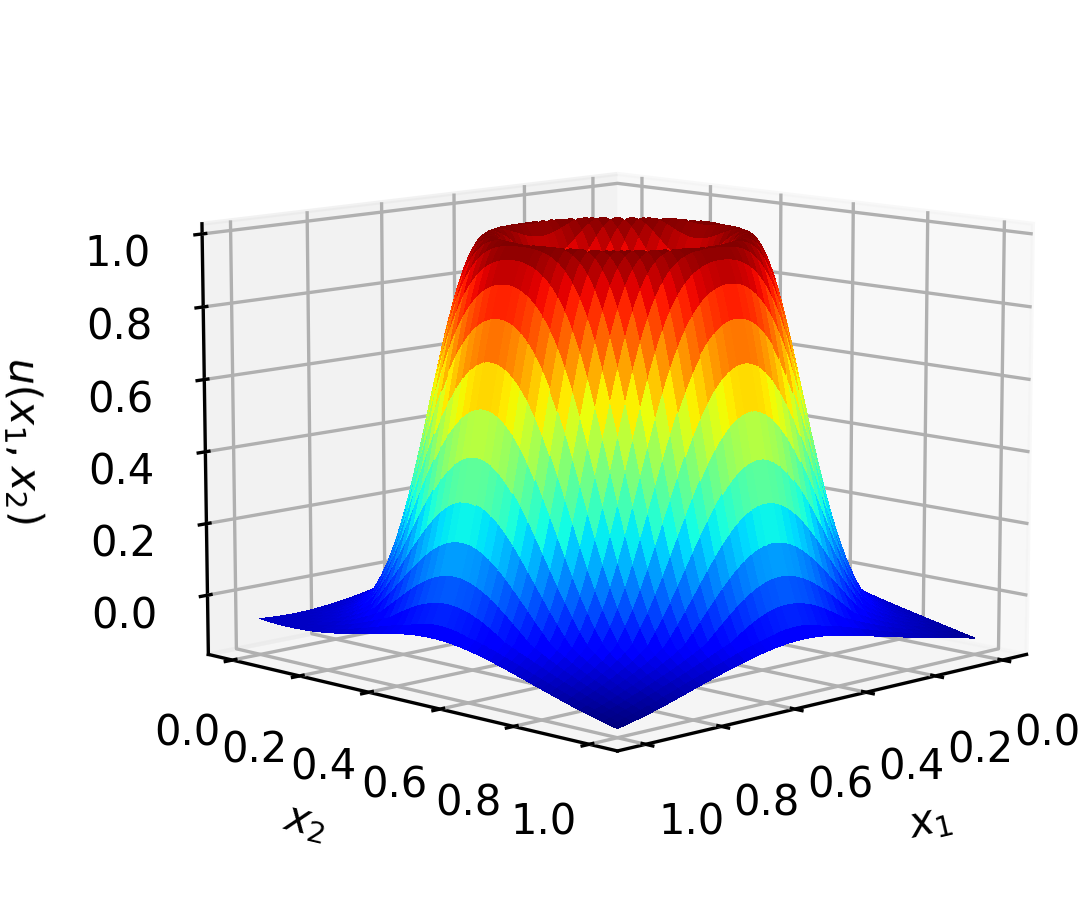}}}\hfill
\caption{Comparison between our approach and PINN  for rotational flow \eqref{2drf} under various diffusion coefficients.}
\label{figurerf}
\end{figure}

Figure \ref{figurerf} shows that our method yields satisfactory predictions,
while the results of PINN have unreasonably negative values near the boundary corners.

\subsection{L-shaped domain}
Consider the following convection-diffusion-reaction problem on a L-shaped domain \cite{Ludwig2016}:
\begin{align}
\label{2dL}
\begin{split}
     -\epsilon \Delta u + \bm{b} \cdot \nabla u  +(3+\sin(2\pi x_1 x_2) ) u &=1-(x_1+x_2)/2,\quad  \bm{x}\in \Omega,
     \\[5pt]
      u&=0,\quad  \bm{x} \in \partial \Omega,
\end{split}
\end{align}
where $ \Omega=(-1,1)^2/(-1,0)^2 $, and $ \bm{b}=-(1+1/2 \sin(2\pi x_1),2-\cos(2 \pi x_2))^T$,
which results in boundary layers occurring at $x_1=0, -1 $ and $ x_2=0,-1 $.

\begin{figure}[!hbt]
\centering
\subfloat[Our prediction ($\epsilon = 10^{-3}$)]{\label{fig:mdleft}{\includegraphics[width=0.33\textwidth]{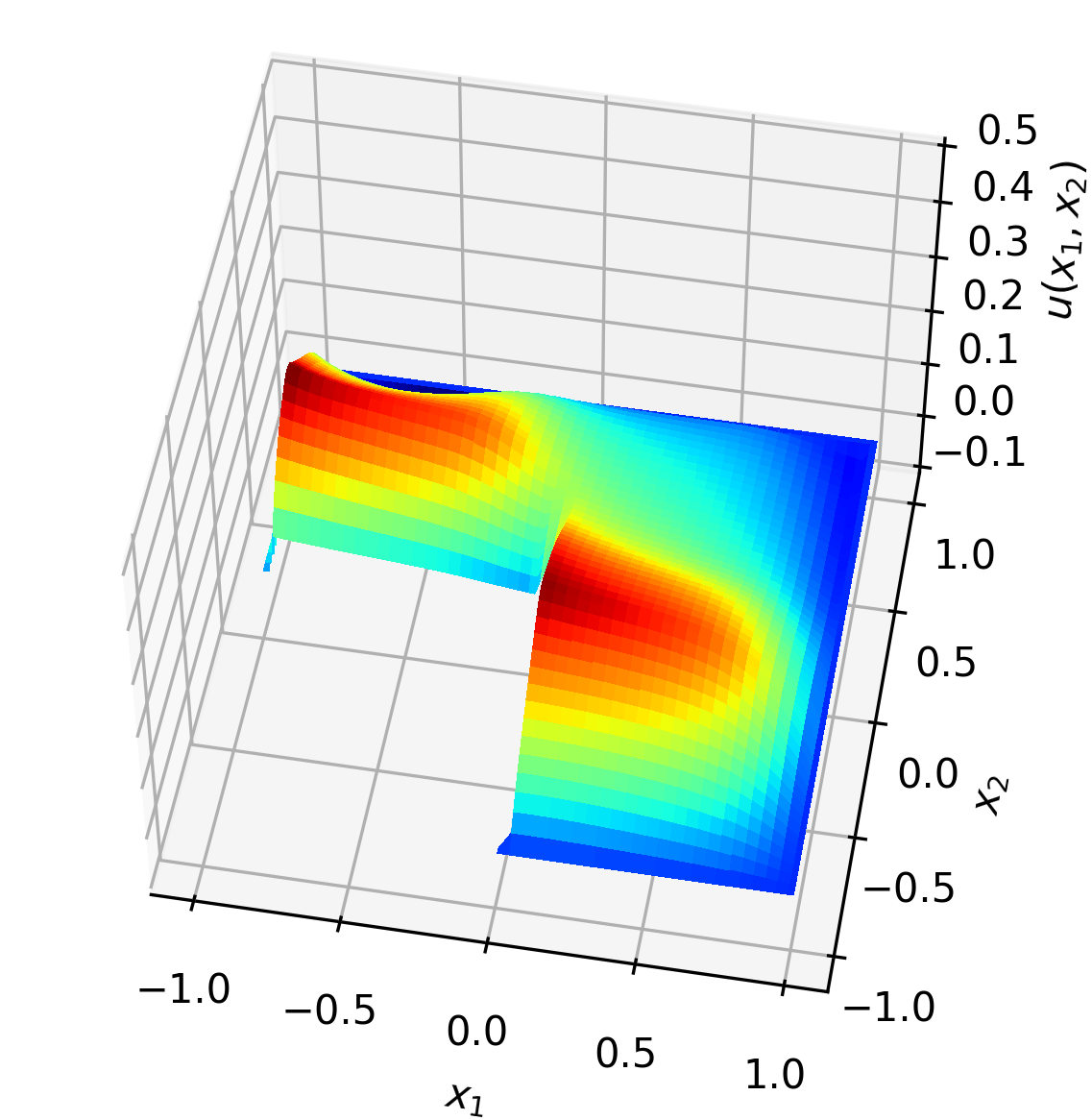}}}\hfill
\subfloat[Our prediction ($\epsilon = 10^{-6}$)]{\label{fig:mdleft}{\includegraphics[width=0.33\textwidth]{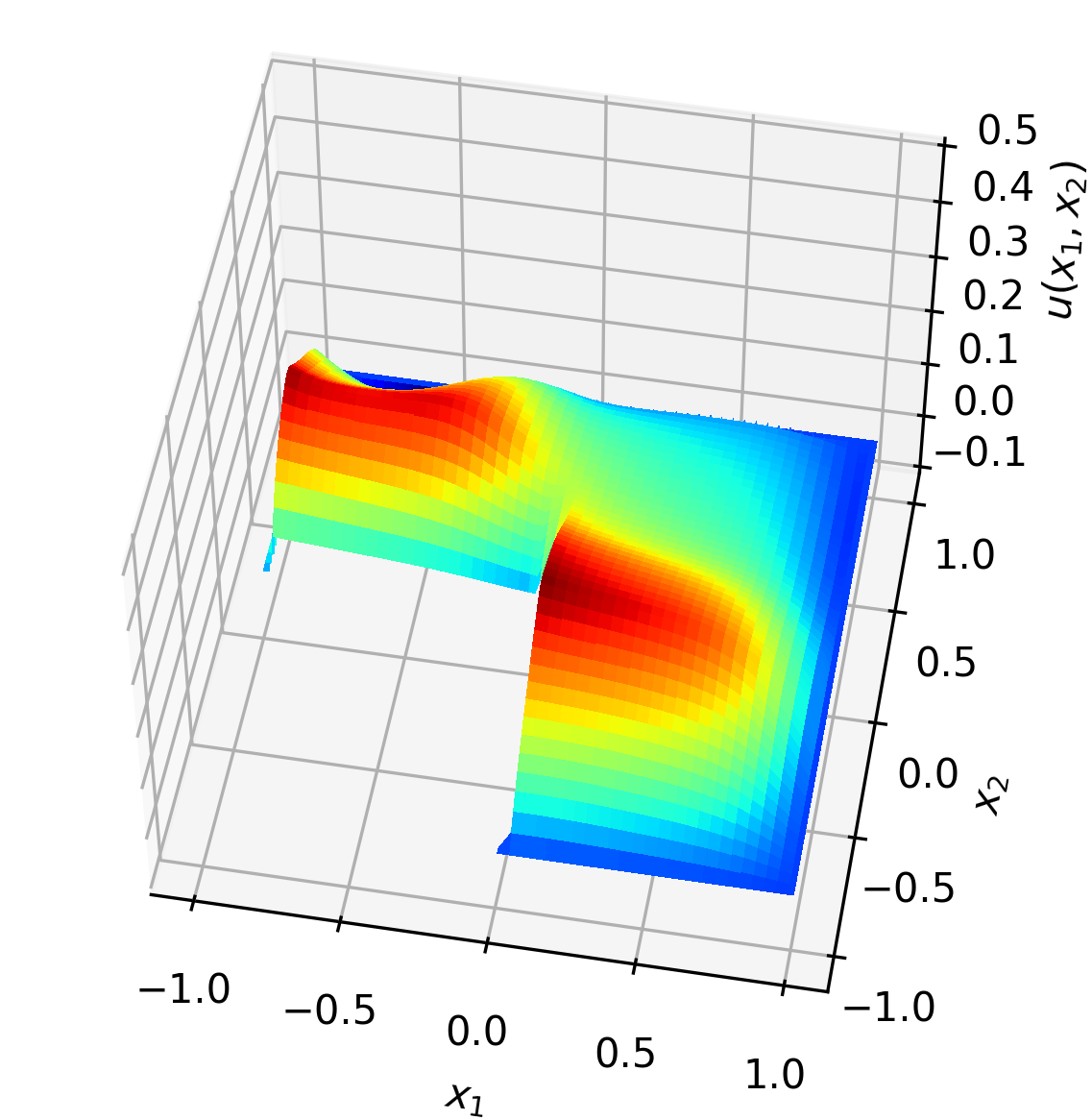}}}\hfill
\subfloat[Our prediction ($\epsilon = 10^{-9}$)]{\label{fig:mdleft}{\includegraphics[width=0.33\textwidth]{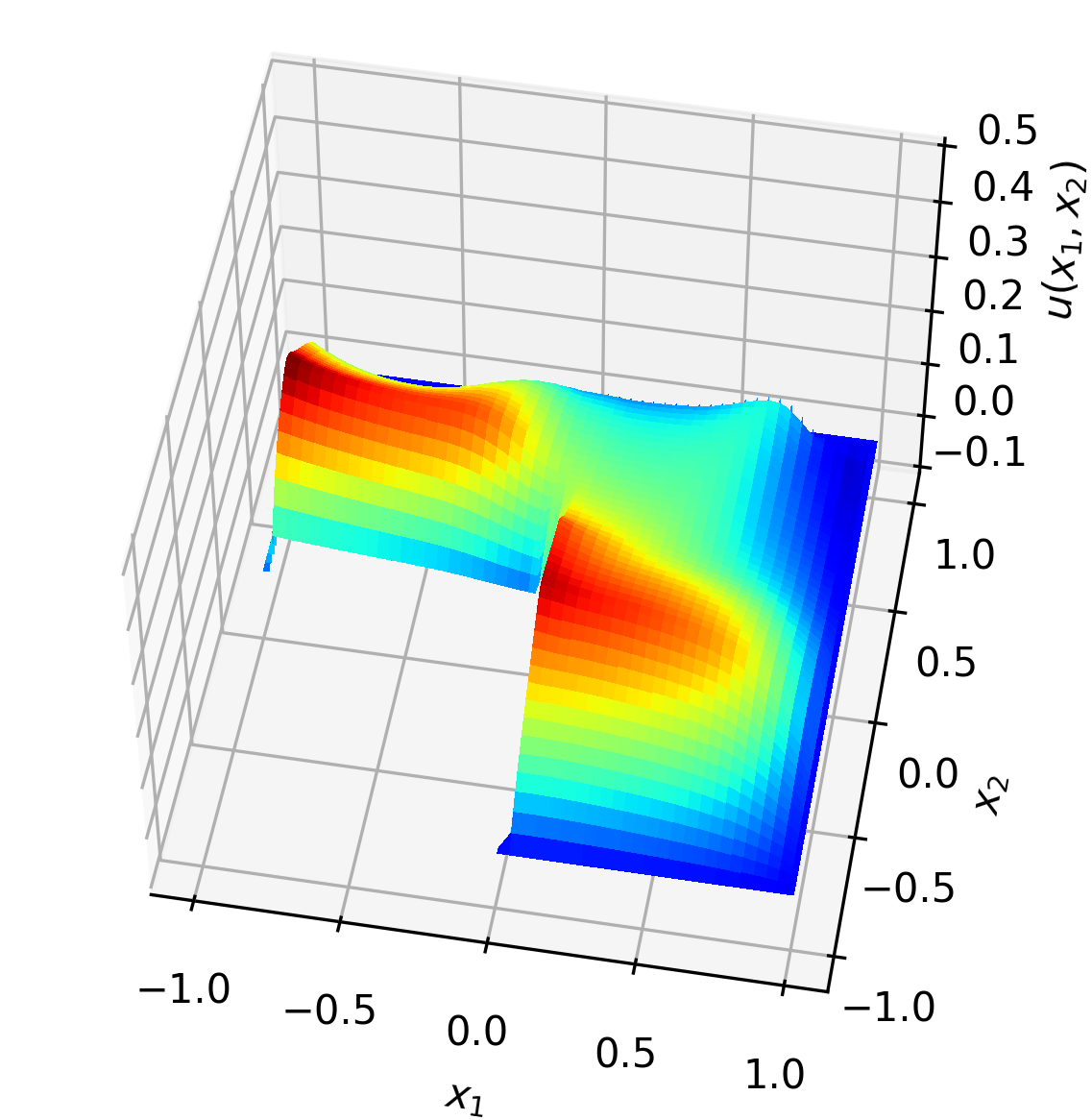}}}\hfill
\subfloat[PINN prediction ($\epsilon = 10^{-3}$)]{\label{fig:mdleft}{\includegraphics[width=0.33\textwidth]{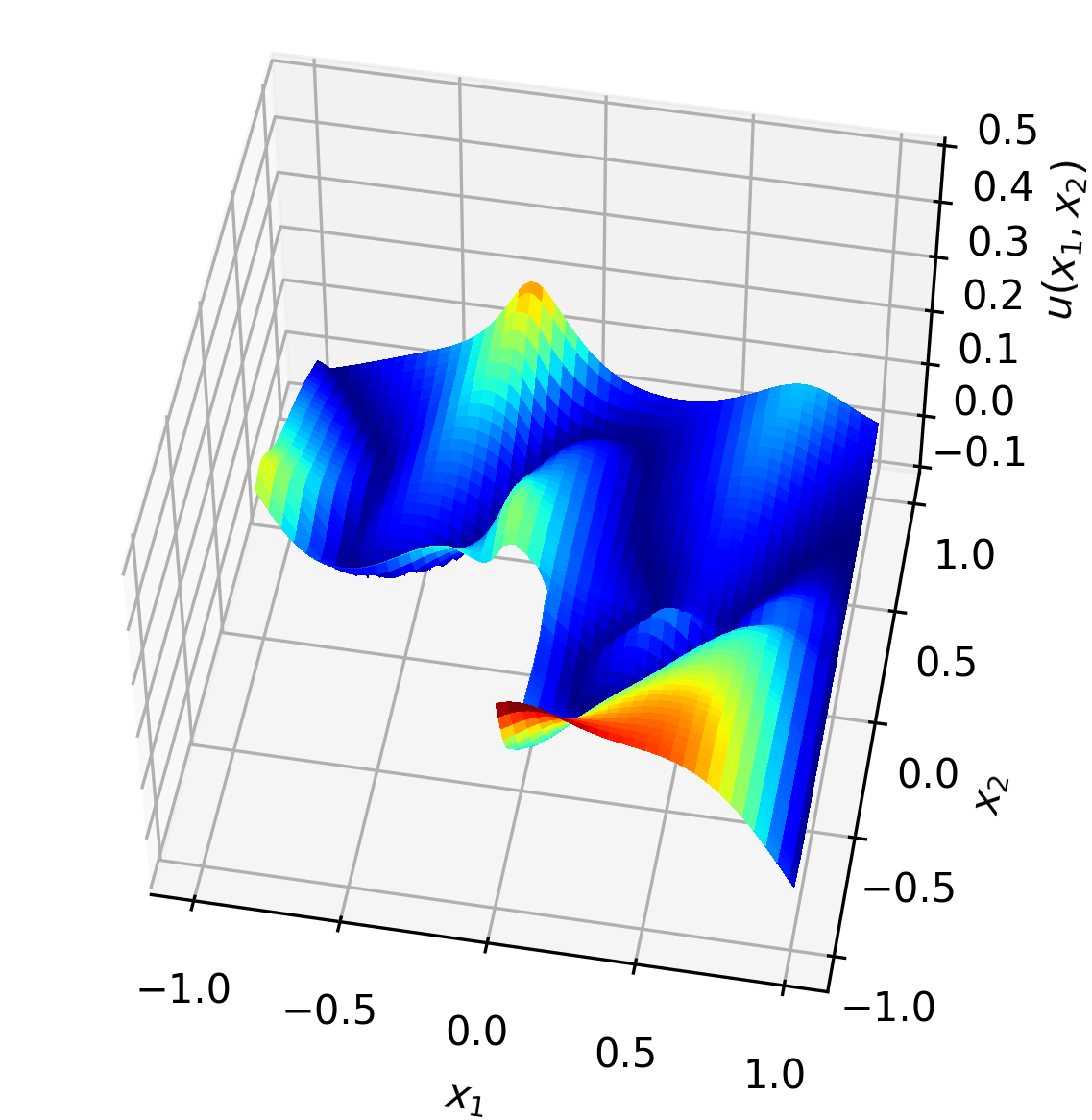}}}\hfill
\subfloat[PINN prediction ($\epsilon = 10^{-6}$)]{\label{fig:mdleft}{\includegraphics[width=0.33\textwidth]{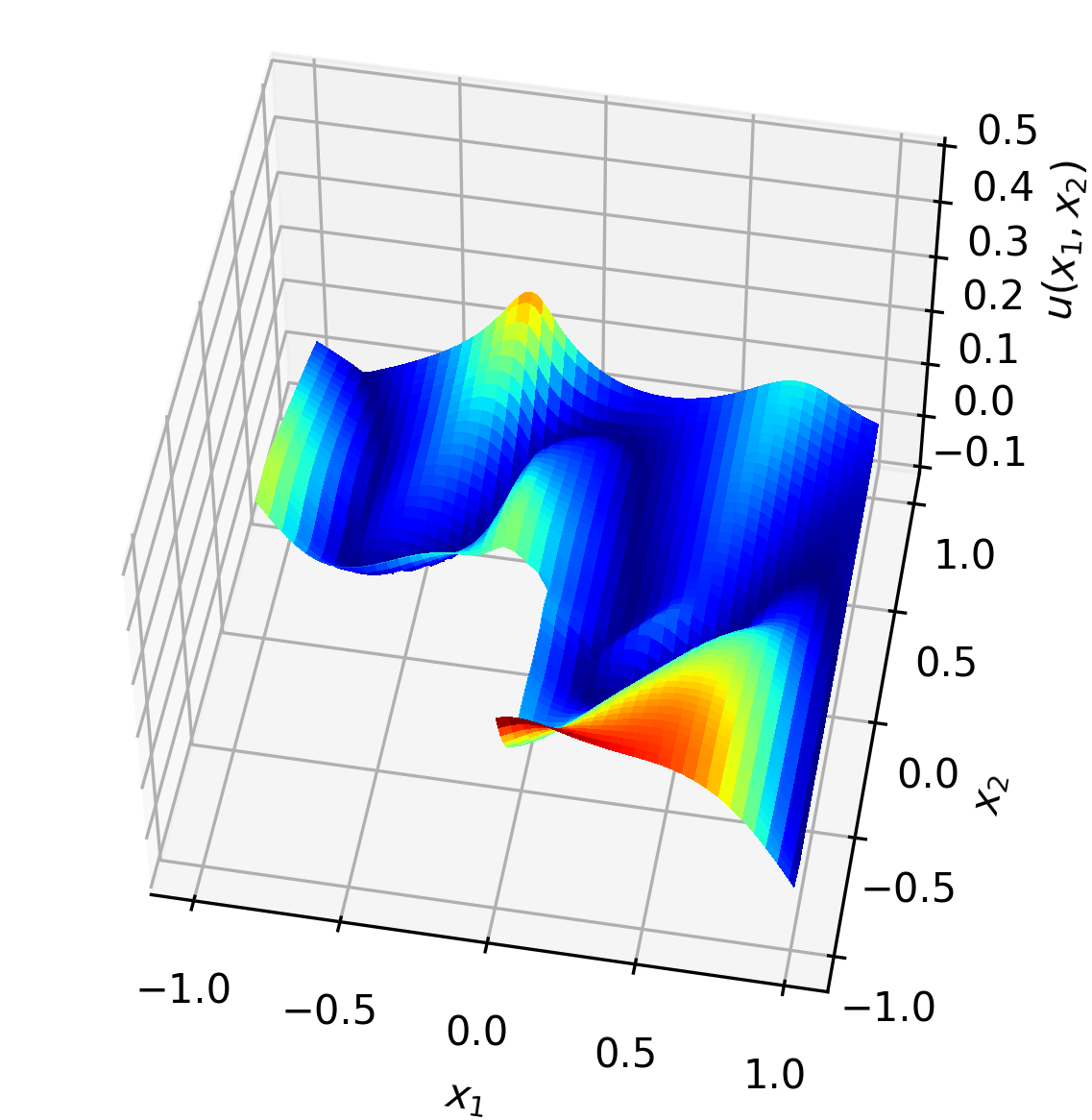}}}\hfill
\subfloat[PINN prediction ($\epsilon = 10^{-9}$)]{\label{fig:mdleft}{\includegraphics[width=0.33\textwidth]{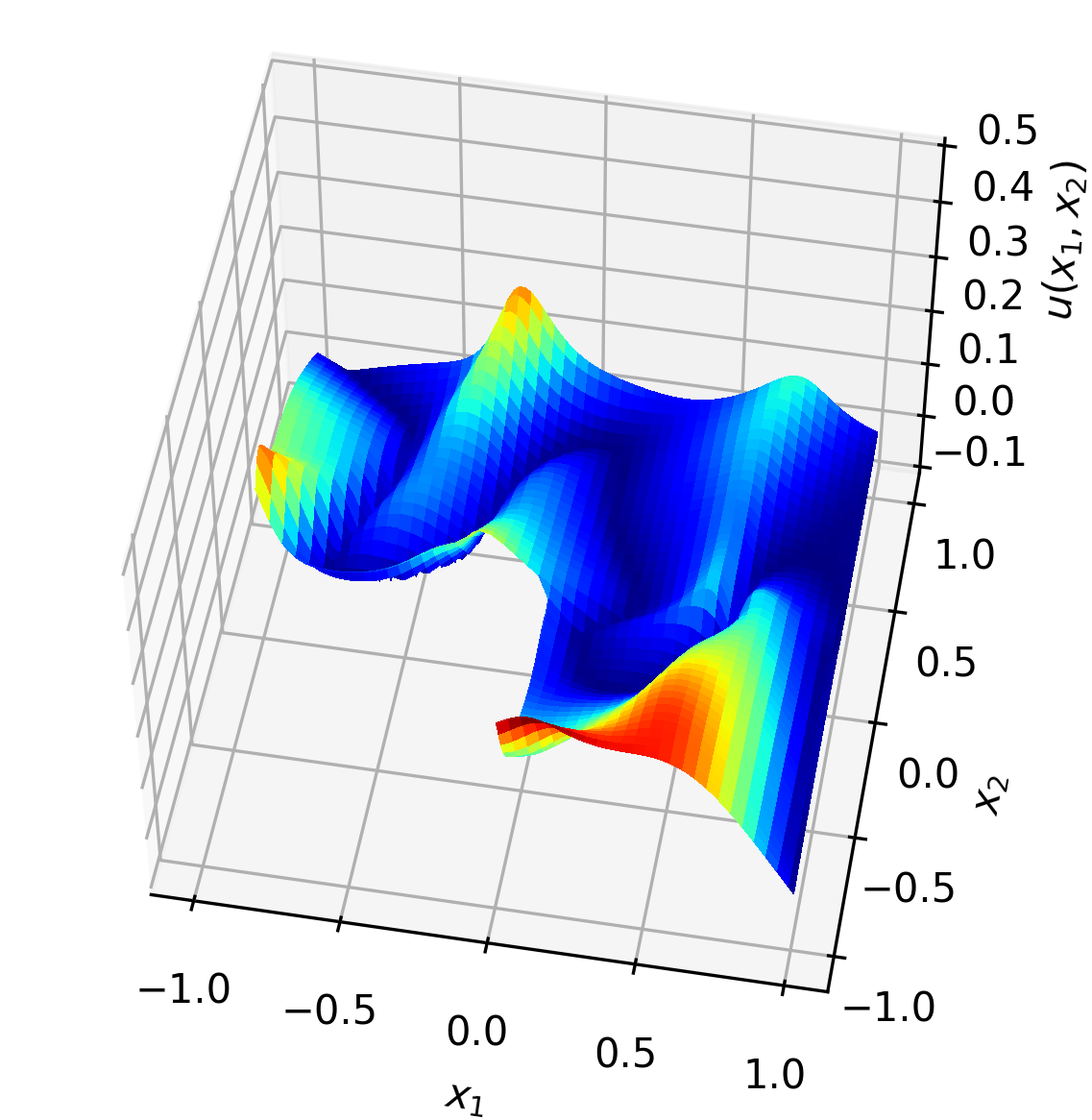}}}\hfill
\caption{Comparison between our approach and PINN  for L-shaped domain equation \eqref{2dL} under various diffusion coefficients.}
\label{figureL}
\end{figure}

From Figure \eqref{figureL},
we can find that the boundary layers are well captured by our approach, with very slight overshooting/undershooting.
In contrast, common PINN performs very poorly and its predictions are highly oscillatory.
It is also noticeable that in this example, our method seems to  slightly degrade in performance as $ \epsilon $ gets smaller.

\subsection{3d singularly perturbed convection-diffusion problem}
For traditional numerical methods, the singular perturbation equations in three spatial dimensions are difficult to solve
due to the huge computational cost.
On the contrary, neural network are more powerful in dealing with high-dimensional problems.
To this end, consider the following three-dimensional convection-diffusion problem:
\begin{align}
\label{3d}
\begin{split}
     -\epsilon \Delta u +\bm{b} \cdot \nabla u  &=f,\quad  \bm{x}\in \Omega,\\
       u&=0,\quad  \bm{x} \in \partial \Omega,
\end{split}
\end{align}
where $\Omega=(0,1)^3$,
$\bm{b}=[1,2,1]^T$,
and  $f(x)$ is chosen such that the exact solution is given by
\begin{align*}
   u=\sin (x_1) (1-e^{-(1-x_1)/\epsilon})(1-x_2)^2 (1-e^{-x_2/\epsilon}) (1-x_3) (1-e^{-x_3/\epsilon}).
\end{align*}
The solution of \eqref{3d} has three exponential layers at $x_1=1 $, $x_2= 0 $ and $ x_3=0$, respectively.

\begin{table}[!htb]
    \center
    \caption{Normalized root mean squared error and computational time of our approach for equation \eqref{3d}.}
    \label{table3d}
    \scalebox{1}{
        \begin{tabular}{ccc}
            \hline
            Diffusion coefficient        & Ours                                 &  PINN           \\
            \hline
            $\epsilon= 1e-3 $           &$4.37 \times 10^{-3}$     &$2.58 \times 10^{-1}$           \\
            \hline
            $\epsilon= 1e-6 $           &$4.46 \times 10^{-3}$     &$4.68 \times 10^{-1}$           \\
            \hline
            $\epsilon= 1e-9 $           &$4.43 \times 10^{-3}$     &$4.72 \times 10^{-1}$           \\
            \hline
    \end{tabular}}
\end{table}

We compare  the errors of our approach with PINN for solving a three-dimensional equation \eqref{3d}.
As can be seen from Table \ref{table3d},
our method obtains about two orders of magnitude lower error than the normal PINN under various $ \epsilon $.

\subsection{Sensitivity analysis}
In our approach, there is an important hyperparameter $G$,
which is used to quantify the magnitude of the gradients of the samples in the subset $ X_{sub} $,
and further helps to determine the threshold $\beta(t) $ for reweighting.
In this subsection, we will study the effect of this hyperparameter.

To this end, we take the equation \eqref{1de} with $ \epsilon=1e-9 $ as an example,
and then apply our approach using $ G=1, 10, 20 $ and $ 30 $, respectively.

\begin{figure}[!hbt]
\centering
\subfloat[Training loss curves]{\label{fig:mdleft}{\includegraphics[width=0.4\textwidth]{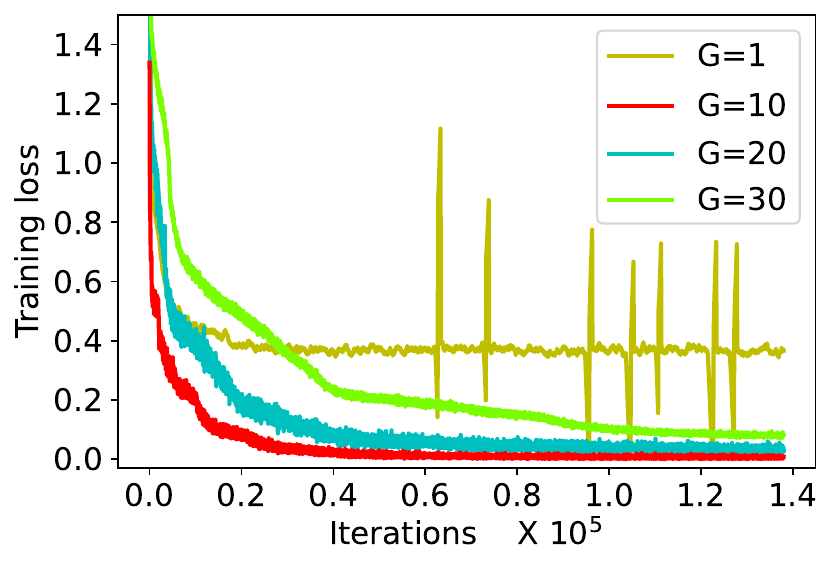}}}\hfill
\subfloat[Predictions]{\label{fig:mdleft}{\includegraphics[width=0.4\textwidth]{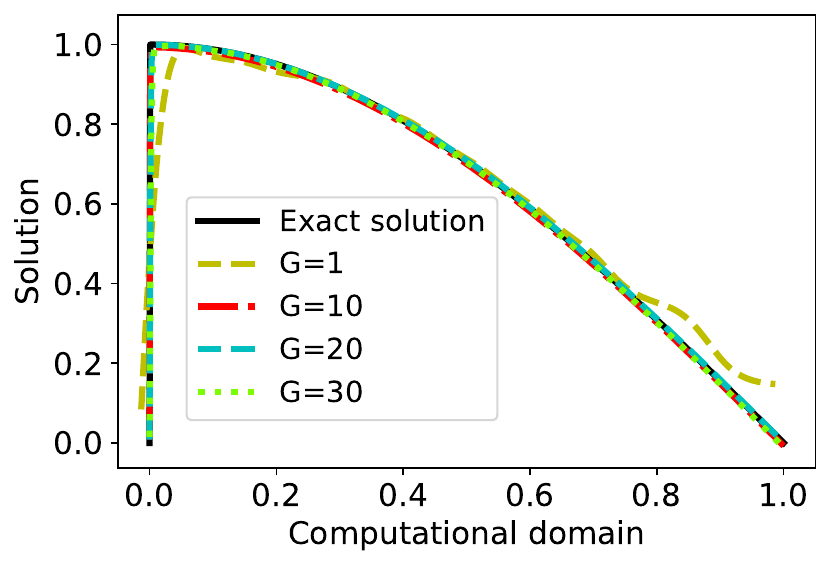}}}\hfill
\caption{Sensitivity studies for the hyperparameter $ G $.}
\label{figureG}
\end{figure}

From Figure \ref{figureG},
we can find that the approach is stable with respect to a large parameter $G$.
Whether $ G=10,20 $ or 30, the training process converges well,
and the corresponding predictions are almost identical.
On the contrary,  $G=1 $ results in an unstable training and larger prediction errors,
which implies that a too small $ G $ cannot well distinguish the gradients from layer or non-layer regions.

\section{Conclusion}
\label{con}
PINNs fail to learn accurate approximations when dealing with singularly perturbed convection-diffusion-reaction problems whose solutions contain sharp  boundary/interior layers.
We studied this failure mode from a regional distribution perspective and revealed that the network fails to converge
due to the extreme multiscale discrepancy in the underlying solutions between regions.
We demonstrated that the widely used approach that prioritizing high-loss regions does not help in training.
A curriculum learning approach was then developed
that emphasizes learning of easier non-layer regions,
thereby significantly improving the prediction accuracy of PINNs.
Our study indicates for the first time that paying less attention to high-loss regions can be a feasible strategy
for accurately learning the equations with strong multiscale characteristics.

\section*{Acknowledgments}
This research is partially supported by National Natural Science Foundation of China (11771257), Natural Science Foundation of Shandong Province (ZR2021MA010) and Graduate lnnovation Foundation of Yantai University (through grant no. GGIFYTU2214).

\end{document}